\pdfoutput=1
\documentclass[11pt]{article}

\usepackage[]{EMNLP2023}

\usepackage{times}
\usepackage{latexsym}
\usepackage{multirow}
\usepackage{multicol}
\usepackage{booktabs}
\usepackage{amsmath}
\usepackage{amsfonts}
\usepackage{pifont}
\usepackage{makecell}
\newcommand{\xmark}{\ding{55}}%
\usepackage{cleveref}
\newcommand{\Scref}[1]{\S\ref{#1}}
\newcommand\ours{\textsc{OLIVE}} 
\newcommand\oursR{\textsc{OLIVE-R}}
\newcommand\oursG{\textsc{OLIVE-G}}
\newcommand\oursRG{\textsc{OLIVE-RG}}
\usepackage{enumitem}

\usepackage[T1]{fontenc}
\usepackage[utf8]{inputenc}

\usepackage{microtype}

\usepackage{inconsolata}
\usepackage{graphicx}
\usepackage{kky}

\newcommand{\draftonly}[1]{#1} 
\renewcommand{\draftonly}[1]{}
\title{OLIVE: Object Level In-Context Visual Embeddings}

\author{Timothy Ossowski$^1$, Junjie Hu$^{1,2}$\\
  $^1$Department of Computer Science, $^2$Department of Biostatistics and Medical Informatics\\
  University of Wisconsin, Madison, WI, USA\\
  \texttt{ossowski@wisc.edu}, \texttt{junjie.hu@wisc.edu}} 

\begin{document}
\maketitle

\begin{abstract}
    Recent generalist vision-language models (VLMs) have demonstrated impressive reasoning capabilities across diverse multimodal tasks. However, these models still struggle with fine-grained object level understanding and grounding. In terms of modeling, existing VLMs implicitly align text tokens with image patch tokens, which is ineffective for embedding alignment at the same granularity and inevitably introduces noisy spurious background features. Additionally, these models struggle when generalizing to unseen visual concepts and may not be reliable for domain-specific tasks without further fine-tuning. To address these limitations, we propose a novel method to prompt large language models with in-context visual object vectors, thereby enabling controllable object level reasoning. This eliminates the necessity of fusing a lengthy array of image patch features and significantly speeds up training. Furthermore, we propose region-level retrieval using our object representations, facilitating rapid adaptation to new objects without additional training. Our experiments reveal that our method achieves competitive referring object classification and captioning performance, while also offering zero-shot generalization and robustness to visually challenging contexts.\footnote{Our code and models are available at \url{https://github.com/tossowski/OLIVE}}

    % \jh{We should be a bit more clear about the challenges: 1) existing VLMs (GPT-4) are still weak at fine-grained object-level understanding and grounding; 2) rare or unseen objects even further exacerbate the problem; 3) in terms of modeling, existing VLMs implicitly align image patch tokens with text tokens which are not effective at the same granularity, and inevitably involve noisy spurious background features. This urges us to have explicit modeling of object-level features to provide accurate, and more robust interpretation of objects -- we need the robustness experiment to support this argument.}

    % Conversely, embedding-based methods such as RegionCLIP and GLIP have the ability to generalize to novel objects in a zero-shot manner, but cannot be applied to downstream generative tasks without training a decoder that accepts these features as input.
\end{abstract}
\section{Introduction}
% \jh{Outline: 1. most VLMs handle the whole image for understanding, leading to two major shortcomings: a) image patch tokens and text tokens are not at the same granularity which creates difficulty for alignment and grounding; b) inefficiency in encoding long-contexts for LLM decoder. 2. (new paragraph) recent region-based VLMs are pre-trained to integrate object level information: GPT4ROI does XXX, Shikra does YYY, etc. Although they provide significant improvement to region-based regions, they still (a) fail at recognizing unseen/rare objects and (b) are sensitive to background/spurious features. Even GPT4V is also not robust to rare objects. 3. (new paragraph) To address the generalization to unseen objects, one straightforward way is to integrate a retrieval-component, motivated by REVEAL and MuRAG which do xxx. Nevertheless, they don't consider object level retrieval and in-context prediction. 4. To address the above issues, we propose object level in-context visual embeddings.}

Despite the popularity, many existing VLMs such as LLaVA \cite{liu2023visual}, MiniGPT4 \cite{zhu2023minigpt}, and mPLUG-OWL \cite{ye2023mplugowl} handle the entire image for visual understanding, leading to two major shortcomings. First, these VLMs use a visual transformer to split an image into a grid of image patches and embed them into a lengthy array of image patch embeddings that have object level features scattered around different positions of the array. This leads to the different granularity between the image patch tokens and text tokens, further creating difficulty in aligning and grounding visual objects to text concepts. Second, feeding all image patch embeddings to the large language model (LLM) decoder is problematic due to the resulting long context and inefficiency of including in-context examples from multiple images.

To improve fine-grained visual alignment, recent region-based VLMs are pre-trained to integrate object level information into the LLM decoder. GPT4ROI \cite{zhang2023gpt4roi} pre-trains LLMs to understand ROIAlign features~\cite{he2017mask} extracted from bounding boxes. Other similar methods such as Shikra \cite{chen2023shikra} or Kosmos-2 \cite{peng2023kosmos} ground and refer to objects using text in multimodal referential dialogues. FERRET \cite{you2023ferret} and ViP-LLaVA \cite{cai2023making} further support free-form shapes as referring input by summarizing visual features sampled within the region of interest. Although these methods provide improvement to object level reasoning, they still fail at recognizing unseen/rare objects and are sensitive to spurious background features, as shown in \Scref{sec:analysis}. Even powerful closed-source multimodal models such as GPT4V are unreliable to deploy in high-stakes domain-specific situations such as the medical domain \cite{senkaiahliyan2023gpt}.

A straightforward way to handle generalization to unseen visual content is to integrate a retrieval component. Methods such as REVEAL \cite{hu2023reveal} and MuRAG \cite{chen2022murag} provide retrieved multimodal facts as supplementary context to help VLMs generalize to new concepts without further training. However, these models do not consider object level retrieval and in-context prediction. Models such as Flamingo \cite{alayrac2022flamingo} and Qwen-VL \cite{bai2023qwen} allow for in-context examples from multiple images, yet do not support object level retrieval and reasoning.

To address the above issues, we propose to encode object level in-context visual embeddings ($\ours$) to enhance LLMs with region-level reasoning capabilities. Critically, we omit lengthy image patch features and encode visual object embeddings by a lightweight encoder of 20 million parameters, allowing for faster training and direct connection to existing LLMs. This preserves the full functionality of the original LLMs, while also introducing novel multimodal reasoning abilities. Furthermore, our object level retrieval module allows for more precise queries and retrieved information to help the model adapt to domain-specific tasks with limited training data. Our contributions are summarized below and in Table \ref{tab:methods}:
\begin{itemize}[leftmargin=13pt]\itemsep-0.2em
    \item We propose a lightweight object encoder that can be connected to existing LLMs to enable controllable object level multimodal reasoning with free-form input annotations.
    \item Our model omits image patch features and summarizes object features into a single vector, significantly reducing context length for more efficient training and inference, and allowing for in-context examples from multiple images.
    \item We conduct extensive experiments with region-retrieval of object level features and showcase rapid adaptation to unseen visual concepts.
\end{itemize}

\begin{table*}[!htbp]
   \resizebox{\linewidth}{!}
    {
    \centering
    \scriptsize
        \begin{tabular}{cccccc}
        % \hline
        \toprule[1pt]
        % \multirow{2}{*}{Method} & \multicolumn{2}{c}{En$\rightarrow$Ru} & \multicolumn{2}{c}{En$\rightarrow$It} & \multicolumn{2}{c}{En$\rightarrow$ES} & \multicolumn{2}{c}{En$\rightarrow$Fr} & \multicolumn{2}{c}{En$\rightarrow$DE}\\
        Model & Free-form Visual Prompts & Free-form Text prompts & Visual Generalization & Generative Approach & Multi-Image \\ \midrule

        Ferret & \textcolor{blue}{\checkmark} & \textcolor{blue}{\checkmark} & \textcolor{red}{\xmark} &  \textcolor{blue}{\checkmark} & \textcolor{red}{\xmark} \\

        Flamingo & \textcolor{red}{\xmark} & \textcolor{blue}{\checkmark} & \textcolor{blue}{\checkmark} &  \textcolor{blue}{\checkmark} & \textcolor{blue}{\checkmark}\\

        GPT4ROI & \textcolor{red}{\xmark} & \textcolor{blue}{\checkmark} & \textcolor{red}{\xmark} &  \textcolor{blue}{\checkmark} & \textcolor{red}{\xmark} \\

        GLAMM & \textcolor{red}{\xmark} & \textcolor{blue}{\checkmark} & \textcolor{red}{\xmark} &  \textcolor{blue}{\checkmark} & \textcolor{red}{\xmark} \\
        
        RegionCLIP & \textcolor{red}{\xmark} & \textcolor{blue}{\checkmark} & \textcolor{red}{\xmark}&  \textcolor{red}{\xmark} & \textcolor{red}{\xmark} \\

         Llama-Adapter v2 & \textcolor{red}{\xmark} & \textcolor{red}{\xmark} & \textcolor{red}{\xmark} &  \textcolor{blue}{\checkmark} & \textcolor{red}{\xmark} \\
         
        ViP-LLAVA & \textcolor{blue}{\checkmark} & \textcolor{red}{\xmark} & \textcolor{red}{\xmark} &  \textcolor{blue}{\checkmark} & \textcolor{red}{\xmark} \\

        \ours & \textcolor{blue}{\checkmark} & \textcolor{blue}{\checkmark} & \textcolor{blue}{\checkmark} &  \textcolor{blue}{\checkmark} & \textcolor{blue}{\checkmark} \\

        % \cline{3-5}
        % {}& $\rightarrow$  & {$\leftarrow$} & $\rightarrow$ & $\leftarrow$ & $\rightarrow$ & $\leftarrow$ & $\rightarrow$ & $\leftarrow$ & $\rightarrow$ & $\leftarrow$ \\
    
        % & RepsNet \cite{tanwani2022repsnet} & - & - & - & - & - & 81.1 \\
        % & Medical Knowledge Pre-training \cite{chen2022align} & 81.9 & 91.4 & 85.6 & 67.6 & 86.8 & 79.2\\
        
        \bottomrule[1pt]
        \end{tabular}
        }
    \vspace{-3mm}
    \caption{ Comparison of $\ours$ to recent VLMs. To the best of our knowledge, we are the first method to offer visual generalization with in-context prompting, while also allowing for free form annotation. A more comprehensive summary of related studies is in \Scref{sec:related}. }
    \label{tab:methods}

\end{table*}

% \begin{figure}[!t]

 % \includegraphics[width=0.5\textwidth]{image/hallucination.png}

%   \caption{\small Weaknesses of generative VLMs. }

%   \label{fig:weaknesses}
% \end{figure}

% \begin{itemize}
%     \item A linear mapping may distort the original image contents and lose crucial information such as compositionality. \cite{caron2021emerging} show that semantic segmentation is possible by looking at the attention of the $\texttt{CLS}$ token. Can we integrate this information directly rather than through linear mapping?
%     \item Dinov2 uses self-distillation to learn visual features in a self-supervised manner. They show that these features can be applied to different tasks such as depth estimation, semantic segmentation, etc with just a tunable linear layer. Can we use these features for vision language understanding/grounding?
%     \item Visual features from a vision transformer occur at the patch level. This does not allow for fine-grained pixel level information to propagate to the multimodal transformer. Can we use smaller patch size to have richer features as done in \cite{caron2021emerging}?
% \end{itemize}

\label{sec:introduction}
\section{Preliminaries}
\label{sec:preliminary}
% This section describes the basic architecture of a generative VLM and the training procedure.

% \jh{The preliminaries should talk about the existing architectures (e.g., LLAVA? or BLIP or MiniGPT4) to combine a vision encoder with a language decoder. }
\paragraph{Generative VLM Architecture} Recent generative VLMs (e.g., LLaVA, BLIP-2) adopt a similar architecture that connects a pre-trained visual encoder $\phi_v$ and a pre-trained language model decoder $\phi_t$ through a lightweight fusion neural network, denoted as $\phi_c$. Specifically, the fusion module first uses a projection function to map a visual feature $\vb\in\Vcal$ to the text embedding space $\Xcal$ of the language model decoder, and then fuse the visual and text embeddings as input to language model decoder. Formally, given an image $v$ and a text prompt $x$, the decoder takes in the combined feature $\xb$ to autoregressively predict the output $y$. 
\begin{align}
    \xb_t = & \texttt{TxtEmbed}(x; \phi_t) \in \Xcal \\
    \vb =& \texttt{ImgEncoder}(v; \phi_v) \in \Vcal \\
    \xb_v =& \texttt{Project}(\vb; \phi_c) \in \Xcal \\
    \xb = & \texttt{Fuse}(\xb_v, \xb_t; \phi_c) \\ \label{eq:p_vlm}
    p_\text{vlm}(y|v, x) =& \prod_{j=0}^{|y|} p_{\phi_t}(y_j | \xb, y_{<j})
\end{align}

Different from prior fusion modules (e.g., linear projection in LLaVA, gated cross-attention in Flamingo, and Q-former in BLIP-2) that project the whole image features, we propose an object level encoder (\Scref{sec:method:object_encoder}) that captures fine-grained region features and speeds up training and inference.

\paragraph{Visual Instruction Tuning} We adopt a similar visual instruction-tuning approach as \citet{liu2023visual} by fine-tuning parts of the VLM parameters (e.g., $\phi_c$ and/or $\phi_t$) on instruction-following data. The training objective is based on maximum likelihood estimation for next-token predictions given the input image and the text prompt. Different from prior work using pure text prompts, our object encoder and retrieval module (\Scref{sec:method:object_encoder}, \Scref{sec:method:region_retrieval}) enables the usage of code-switched prompt sequence mixing text tokens and image object tokens, and the rapid adaptation to unseen domains via in-context prediction.

%used to maximize the  objective is based on traditional next-token prediction through teacher forcing. 
% Our experiments involve a mix of retrieval-based non-parametric methods which require no training in addition to finetuning based approaches. For the latter case, the training objective is based on traditional next word prediction loss for GPT training. Formally, given model parameterized by $\phi$ with input text and object level features $\mathbf{X}$ and ground truth answer string $z$, we formulate the likelihood of the prediction as:
% \begin{align}
%     P_\text{gen}(z | \mathbf{X}) = \prod_{j=0}^{|z|}P_{\phi}(z_j | \mathbf{X}, z_{<j}).
% \end{align}
% Our models are trained based on a cross-entropy loss between the ground-truth label and this conditional probability likelihood.

\section{Method}
\label{sec:method}
\begin{figure*}[th!]
  \centering{\includegraphics[width=\textwidth]{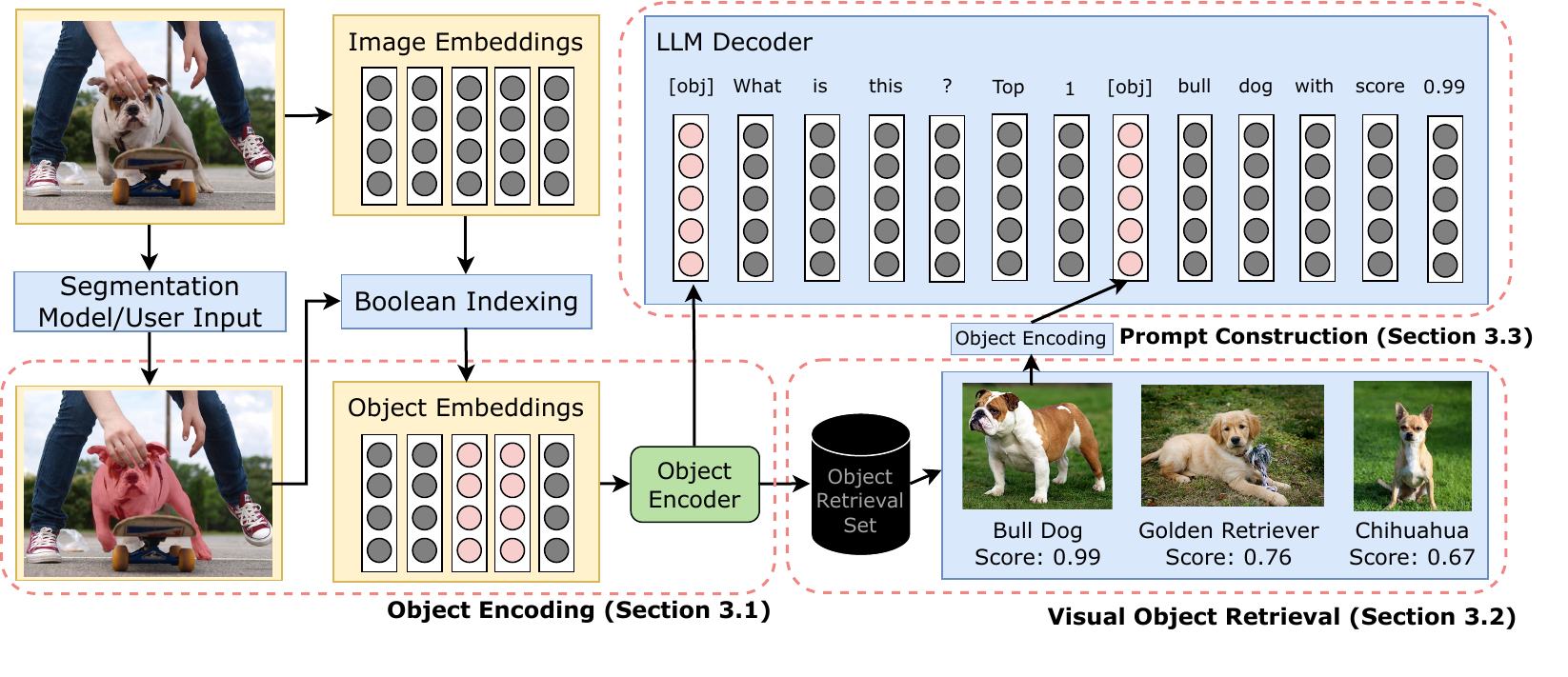}}
  \vspace{-8mm}
  \caption{An overview of our method which consists of three main components, described in detail in Section \ref{sec:method}. The object encoder (green) is the only module with required trainable parameters. Note: the prompt in the LLM decoder is modified slightly for visual clarity. The exact prompt can be found in Appendix \ref{sec:prompts}.}
  \label{fig:prompt}
  \vspace{-3mm}
\end{figure*}
 
This section as well as Figure~\ref{fig:prompt} highlights the main components of our method. We first design an object encoder (\Scref{sec:method:object_encoder}) to learn visual object embeddings in a shared vision-text space, then apply a similarity search over object embeddings to retrieve relevant visual objects (\Scref{sec:method:region_retrieval}), and finally construct a code-switch multimodal prompt to integrate the retrieved object information for generation (\Scref{sec:method:prompt_construction}). %The main components of our model are depicted in Figure \ref{fig:prompt}.

\subsection{Object Encoder}
\label{sec:method:object_encoder}

Following popular region-grounded models such as FERRET~\cite{you2023ferret}, we allow for free-form annotation of objects using the object segmentation mask $\ob_\text{mask}$ as input. Specifically, we first encode an image $v$ with a vision transformer \cite{dosovitskiy2020image} to obtain patch-level features $\mathbf{v}$:
\begin{align} \label{eq:patch}
    \mathbf{v} &= \texttt{ImgEncoder}(v; \phi_v) \in \mathbb{R}^{(n^2 + 1)\times d},
\end{align}
where $n$ is the grid size and $d$ is the dimension of hidden states. To further obtain an object level feature $\mathbf{v}_{\text{obj}}$ from the image, we first extract a subset of the image features $\mathbf{v}_\text{masked}$ corresponding to the binary object segmentation mask $\mathbf{o}_{\text{mask}}$:
\begin{align} \label{eq:v_masked}
    \mathbf{v}_\text{masked} &=  \mathbf{v}[\texttt{Flatten}(\mathbf{o}_{\text{mask}})]\in \mathbb{R}^{l \times d}
\end{align}
where $\ob_\text{mask}$ is a $n\times n$ binary matrix, indicating the corresponding image patches occupied by an object in the image, and $l$ denotes the number of the occupied patches. These segmentation masks can be created by automatic segmentation tools such as SAM~\cite{kirillov2023segment} or provided by human selection on the image. The segmentation mask is first flattened and used to select object patches $\vb_\text{masked}$ from $\vb$. Finally, we obtain the object embedding by compressing $\mathbf{v}_{\text{masked}}$ into a single vector $\mathbf{v}_{\text{obj}}$.% \jh{why do you cite these works here?} \tim{These are other works which compress the entire image into a few vectors (e.g. BLIP, Flamingo)}

\begin{align} \label{eq:obj_encoder}
    \mathbf{v}_{\text{obj}}  &=  \texttt{ObjectEncoder}(\mathbf{v}_\text{masked}; \phi_c)\in \mathbb{R}^{d},
\end{align}
where the object encoder uses a lightweight 2-layer transformer that acts similar to a visual resampler~\cite{zou2023generalized, li2023blip}, followed by a learnable linear layer to further project the visual representation to the text space \cite{liu2023visual}.

\subsection{Visual Object Retrieval}
\label{sec:method:region_retrieval}
In many cases, the object of interest does not resemble anything seen during training. With our visual object embeddings, we can easily perform object level retrieval to match an open class of visual objects and integrate the retrieved information into the language decoder for predicting unseen or rare objects from specific domains (e.g., biomedicine). To this end, we assume access to a retrieval set $\mathcal{R}=\{(\ob_i, d_i, v_i)\}^{m}_{i=1}$, where each triple consists of an object's segmentation mask $\ob_i$, the object's text description $d_i$ and the image $v_i$ containing this object. To retrieve relevant objects from $\Rcal$, we use a similar object encoding as \Scref{sec:method:object_encoder} except that we use the mean pooling of $\vb_\text{masked}$ as the object encoder in Eq.~(\ref{eq:obj_meanpool}), since this simple strategy does not require any learnable parameters for projection to the text embedding space and visual object embeddings can be pre-computed before any fine-tuning. However, we use a learnable object encoder in Eq.~(\ref{eq:obj_encoder}) to connect object embeddings to the LM decoder during instruction-tuning for text generation (\Scref{sec:method:prompt_construction}).
\begin{align}\label{eq:obj_meanpool}
    \mathbf{v}_{\text{obj}}  &=  \texttt{MeanPool}(\mathbf{v}_\text{masked})\in \mathbb{R}^{d},
\end{align}
%\footnote{We use the mean pooling of $\vb_\text{masked}$ as the object encoder for retrieval, since this simple strategy does not require any learnable parameters for projection to the text embedding space and works well. However, we use the learnable 2-layer transformer and a linear layer as the object encoder to connect object embeddings to the LM decoder for text generation.} 
During retrieval, we compute a query vector $\mathbf{v}_{\text{query}}$ for a given object, and compute the cosine similarity scores between $\mathbf{v}_{\text{query}}$ and all the visual object embeddings from $\Rcal$ to obtain the top $k$ closest triples, denoted as $\Kcal = \{(\ob_i, d_i, v_i)\}_{i=1}^k$. %\jh{when you get the retrieved objects here in $\Kcal$, are you saving their visual embeddings computed by Eq.~\ref{eq:v_masked} and reuse these embeddings during fine-tuning? or you're saving the objects' segmentation masks and computing their object embeddings on the fly using Eq.~\ref{eq:obj_encoder} during fine-tuning?} \tim{It is the latter. The meanpool vector is only used for retrieval. Once the we get the retrieved objects, we use the object encoder to get the object representation for the model.}%by averaging and normalizing the image features $\mathbf{v}_\text{masked}$ from Eq. (\ref{eq:v_masked}).  %\jh{You are actually using the $\text{MeanPool}(\vb_\text{masked})$?} \tim{What do you mean by MeanPool? Right now, I average the patch features and normalize to get the retrieval query vector. I think it can be improved with either crop/learned embedding. But this works for adaptation already. I think after submission we can try using the learned object encoder from object detection task -> create retrieval dataset} \jh{$\vb_\text{masked}$ is a $l\times d$ matrix. I think you are using the mean pooling of $\vb_\text{masked}$ as the vector representation.} \tim{Basically taking torch.mean (across the l dimension)} \jh{I got the point -- it's a notation mistake.}%\jh{Why are you using $\vb_\text{masked}$ instead of $\vb_\text{obj}$? Because you want to use the encoder before any fine-tuning?} \tim{Because the object encoder is randomly initialized, it isn't useful for retrieval until it is finetuned. We could do something like REALM, but I wanted to try this simple method first} \jh{Okay. we should explain this in a footnote.} 

% We can then use $\mathbf{o}_{\text{query}}$ to compute the similarity scores between all the similarly constructed object level key embeddings $\mathbf{o}_{\text{key}}$ to retrieve the top $k$ closest (object, document) pairs $\mathcal{K} = \{(o_i, d_i)\}_{i=1}^k$. We use dot product as our similarity function, and maximum inner product search (MIPS) may be used to efficiently find the top-$k$ most relevant entries. 
\subsection{In-context Prompt Construction}
\label{sec:method:prompt_construction}
As the visual object embeddings are projected into the text embedding space of the LM decoder, this allows us to construct a code-switched prompt that mixes visual objects with text tokens for the LM decoder (e.g., Llama 2~\cite{touvron2023llama}). In addition, as our object encoder compresses a visual object into a single vector $\vb_\text{obj}$, this significantly shortens the length of the visual tokens that the LM decoder needs to fuse with text tokens. Therefore, we can easily integrate multiple retrieved object embeddings into the prompt to augment the LM decoder for in-context text generation. Specifically, we define a special vocabulary token \texttt{[obj]} which can be inserted flexibly in the user prompt $x$. For example, the user can ask ``\texttt{[obj]} Describe this part of the image" to perform region-level description. The embedding of this token is directly replaced with its corresponding visual object embedding. Formally, given a text prompt $x$ that contains indexed \texttt{[obj]} tokens referring to an object $\vb_\text{obj}$ of interest in an image $v$ and its relevant objects in $\Kcal$, we define a prompting function that replaces the text embedding of \texttt{[obj]} with its corresponding visual object embedding, and integrates the top $k$ most similar objects $\mathcal{K}$ as in-context examples. For example, a prompt with retrieved in-context examples can be ``The top [k] related objects are: [obj$\_1$] is a [label],...[obj$\_k$] is a [label]. [obj$\_\text{query}$] What is this?''. We provide more details about in-context prompt templates and construction in Appendix \ref{sec:prompts}.
\begin{align} \label{eq:prompt}
    \xb = \texttt{Prompt}(x, \vb_\text{obj}, \Kcal)
\end{align}

% Formally, given the input object feature(s) $\mathbf{o}_i$, and user text query $\mathbf{T}$, our model decodes an answer $\mathcal{A}$ with the following equation:
% \begin{align}
%     \mathcal{A} &= \texttt{LLMDecoder}(\texttt{Format}(\mathbf{T}, \mathbf{o}_i, \mathcal{K}))
% \end{align}
%we  compared to prior VLMs (e.g., LLaVA) that directly fuse the patch-level features $\vb$ of the whole image (Eq.~\ref{eq:patch}) with the object information scattering around different positions in the patch features $\vb$. 
% \jh{When using the retrieved objects in the prompt, I assume you're using the object embeddings computed in Eq. \ref{eq:obj_encoder}?} \tim{Yes that's right}
%where \texttt{Format} is a function that replaces the text embedding of \texttt{[obj]} with its corresponding object feature, and integrates the top $k$ most similar objects $\mathcal{K}$ as in-context examples. 
Finally, we feed the multimodal prompt $\xb$ into the LM decoder for text generation following Eq.~(\ref{eq:p_vlm}).  %Note that we deliberately omit the image patch features from the input to the LLM decoder. 
Note that compared to prior VLMs (e.g., LLaVA) that directly fuse the patch-level features $\vb$ of the whole image (Eq.~\ref{eq:patch}) with object information scattering around different positions in $\vb$, our object encoding is computationally more efficient and speeds up the training that involves multiple in-context objects in the multimodal prompt.%and the object vectors provide a rich enough representation to perform object level reasoning about the image.

\section{Experimental 
Settings}
\label{sec:setting}
In this section, we first describe two main object-level tasks for evaluation (\Scref{sec:exp:tasks}) together with the datasets used (\Scref{sec:exp:dataset}). Finally, we describe three variants of our model (\Scref{sec:exp:variants}), the training details \Scref{sec:exp:training}), and the other baselines in comparison (\Scref{sec:exp:baselines}).

%This section describes the different settings of our method in which we run our experiments: Retrieval Setting (\Scref{sec:method:retrieval_only_setting}), Generative Setting (\Scref{sec:method:object_only_setting}), and Combined Setting (\Scref{sec:method:object_and_retrieval_setting}). We then discuss the training settings and models used (\Scref{sec:method:training_settings}). Finally, we describe the datasets used in our experiments (\Scref{sec:setting:dataset}).

\subsection{Object-level Tasks}
\label{sec:exp:tasks}
% \jh{I don't think we need to mention the task here. We could move this part to Sec 4}
\paragraph{Referring Object Classification}
Given an object referred by its image location (e.g. segmentation mask/bounding box), the model is instructed to generate a text that predicts the object's class label in a predefined label set, $\mathcal{C} \in \{c_1, c_2, ... c_n\}$. We provide the ground truth segmentation mask to eliminate localization errors and focus on evaluating the models' understanding of image objects. 

% \jh{do you mean the bounding box location?} \tim{We provide the segmentation mask where available. Otherwise, the bounding box location can easily be used as well}

\paragraph{Referring Expression Generation} Given an input image object referred by a segmentation mask, the model is instructed to generate a natural language expression which semantically matches multiple ground-truth references $\mathcal{R} \in \{r_1, r_2, ... r_m\}$. We use METEOR \cite{banerjee2005meteor} and CIDEr \cite{vedantam2015cider} score for evaluating generated description quality.

\subsection{Datasets}
\label{sec:exp:dataset} 
This section describes the different datasets used in our experiments, with more details in Appendix \ref{sec:datasets}. 
% \jh{shorten this part}
\paragraph{Common Objects in Context (COCO)}~\cite{lin2014microsoft} is a popular visual reasoning dataset with over 800,000 object-level annotations for 80 categories of objects. We use it to train our model to understand region input since it contains high-quality segmentation annotations. We use the standardized train and validation 2017 splits for the detection task, and discard a few ($<$1\%) small segmentation annotations that fail to be converted into a binary mask. Following \cite{zhong2022regionclip}, we evaluate in the setting where ground-truth segmentations are provided as input to eliminate localization errors. We use the standard metric of mean average precision (mAP) for object detection using the COCO API,\footnote{\url{https://github.com/cocodataset/cocoapi}} as well as overall accuracy.

\paragraph{refCOCOg}~\cite{kazemzadeh2014referitgame} is a variant of the COCO dataset with about 50,000 annotations for objects and their description. We use the data to train our model to describe image regions and use their standardized train/validation split.

\paragraph{ChestX-Ray8 (CXR8)}~\cite{wang2017chestx} is a medical dataset consisting of 108,948 frontal-view X-ray images. The image annotations for the 8 possible pathologies are text-mined from the radiology reports using NLP tools. A small subset of 984 images contains bounding box annotation of the pathology. We use this subset for our zero-shot domain adaptation experiments, splitting the data into 16\% retrieval set and 84\% test data. The retrieval set consists of 20 examples of each pathology, and we use overall accuracy as the evaluation metric.

\subsection{OLIVE Variants}
\label{sec:exp:variants}
% \paragraph{Retrieval Setting $(\ours_{retrieval})$}
\paragraph{\oursR~(Retrieval-only)}
\label{sec:method:retrieval_only_setting}
This retrieval-only method predicts the answer to the user question by taking a majority vote of the top $k$ retrieved examples. For simplicity, we fix $k = 5$ for this setting unless otherwise specified and analyze the effect of $k$ in Figure \ref{fig:retrieval_set_size}. Although simple, this baseline proves to be effective and provides salient additional context as described in \Scref{sec:method:object_and_retrieval_setting}. However, this discriminative model does not allow for free-form text generation for tasks such as region captioning. 

\paragraph{\oursG~(Generative-only)}
\label{sec:method:object_only_setting}
This model is trained to generate free-form text based solely on the user question and corresponding object features. We omit the retrieved information to observe the capability of the standalone object representations. We find that even without retrieval, the model can learn to perform more challenging object-level tasks such as region description. The final decoder input can be expressed as a variant of Eq. (\ref{eq:prompt}):
\begin{align}
   \xb = \texttt{Prompt}(x, \vb_\text{obj}).
\end{align}

\paragraph{\oursRG~(Full)}
\label{sec:method:object_and_retrieval_setting}
Our full model generates text outputs based on in-context object examples from retrieval. The multimodal in-context prompt is constructed using Eq. (\ref{eq:prompt}). This prompt includes the retrieved object features, their labels, and their similarity scores. The exact construction can be found in Appendix \ref{sec:prompts}. The top $k$ retrieved multimodal documents in $\mathcal{K}$ are obtained using the same retrieval described in~\Scref{sec:method:region_retrieval} and ordered in increasing relevance score. Both $\oursG$ and $\oursRG$ use greedy decoding for text generation.

% Greedy decoding is used to generate the output for both $\oursG$ and $\oursRG$.

\subsection{Training Details}
\label{sec:exp:training}
Our model uses a frozen \texttt{ViT-L/14} vision transformer from a CLIP model to obtain patch-level features. For our LLM backbone, we use either Llama 2-7B or GPT-2 (124M)~\cite{radford2019language}. The LLM is finetuned with LoRA \cite{hu2021lora} as we find this improves model performance. We use the train splits of two different region-level datasets (i.e., COCO, refCOCOg) as our training data for their respective tasks, and evaluate models on their corresponding validation splits because their test data does not have object-level annotation. More details are in Table \ref{tab:hyperparameters} and we leave the other hyperparameter search to future exploration. We additionally find that we can train a multi-task model by combining the datasets for all object-level tasks (Details in Appendix \ref{sec:multi_task_model}).

% Our object encoder is the only trainable module, which we train for 1 epoch with minibatch size 4 on 4 Nvidia GTX 3090 GPUs. The training takes about 1.5 hours for the object captioning task, and 1 day for the object classification task. A learning rate of \texttt{2e-5} is used with Adam optimizer with default parameters. 

\subsection{Other Baselines in Comparison}
\label{sec:exp:baselines}
% \jh{shorten this part}
% Here we describe related works which we compare our method to.
\paragraph{CLIP} \cite{radford2021learning} Contrastive Language Image Pretraining learns a joint vision-language space between images and their matching captions. We use this method for zero shot object classification by predicting the target with the highest cosine similarity to the cropped region. 

% We then apply the standard practice of encoding the query image along with the target classes, and predicting the class with the highest cosine similarity score.

% \paragraph{BioMedCLIP} \cite{zhang2023large} This work introduces the PMC-15 dataset, which contains a large quantity of multimodal medical data extracted from scientific articles. From this data, 
 \paragraph{BioMedCLIP} \cite{zhang2023large} 
The authors train a CLIP model aligned to biomedical image-text pairs, achieving state of the art on a variety of medical tasks. We use this model as a baseline for object classification in the medical domain.

\paragraph{RegionCLIP} \cite{zhong2022regionclip} This model learns region-text level alignment through soft-labels obtained from CLIP. We use it for referring object detection based on ROIAlign features. 

\paragraph{Kosmos 2} \cite{peng2023kosmos} This generative VLM trains a LLM decoder to perform a variety of visual grounding tasks from their newly introduced grounded image-text (GRIT) dataset. We compare with their results on referring expression generation on the refCOCOg dataset.

\paragraph{Flamingo} \cite{alayrac2022flamingo} This generative model learns to connect frozen visual features and LLMs by training on interleaved image-text data. We evaluate Flamingo's few-shot performance on referring expression generation on cropped image regions. We use an open-source implementation trained on the multimodal C4 \cite{zhu2023multimodal} and LAION-2b \cite{schuhmann2022laion} datasets.

\section{Results and Analysis}
\label{sec:analysis}

\begin{figure*}
    \centering
    \includegraphics[width=\textwidth]{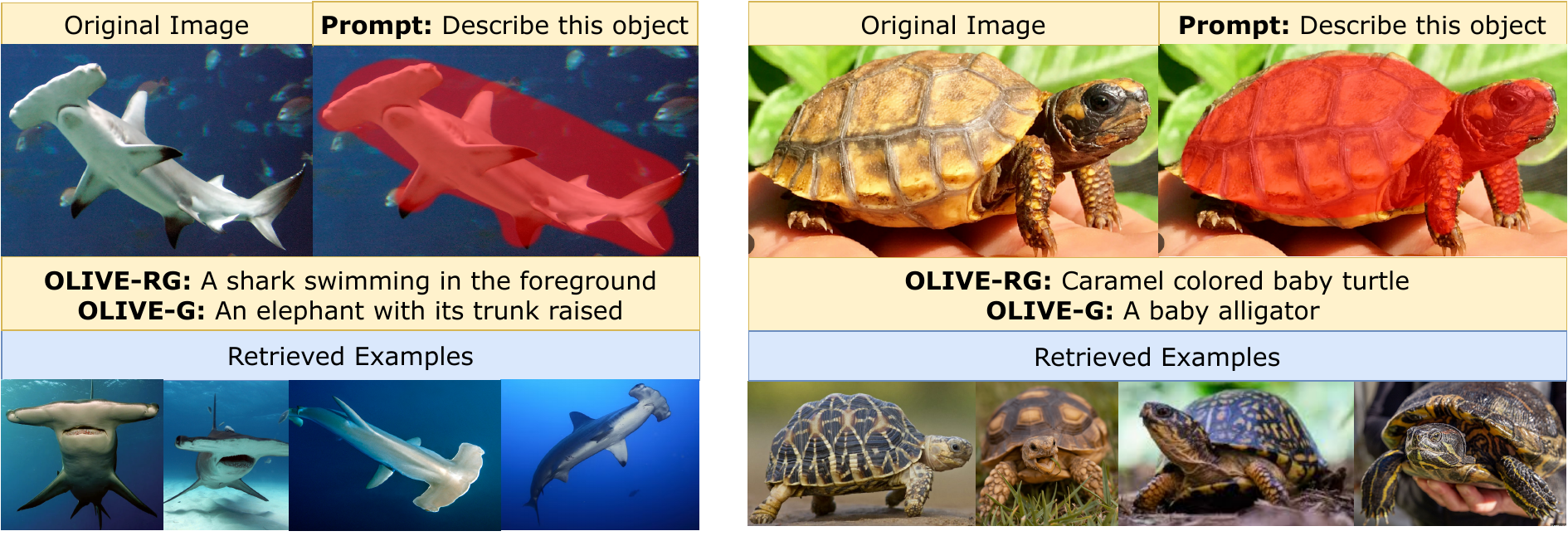}
    \caption{\small Examples showcasing the benefit of using retrieval for out of distribution objects. Despite not being trained with any images of sharks or turtles, $\oursRG$ can describe them zero shot by adding a few pictures of them in the retrieval set.}
    \label{fig:turtle}
\end{figure*}

\begin{table}[htb!]    

    \renewrobustcmd{\bfseries}{\fontseries{b}\selectfont}

    \centering
    \resizebox{\linewidth}{!}
    {
        \begin{tabular}{cllcc}
        % \toprule
        \toprule[1pt]
         Method Type & Pre-Training Data & Method & Accuracy \\

        % Method & {En$\rightarrow$FR} & {En$\rightarrow$KO}\\
        \midrule
        \multirow{6}{*}{Classification} & None & $\oursR$ & 33.5
         \\
         & PMC-15 & BioMedCLIP & 32.5 \\
         & PMC-15 & $\text{BioMedCLIP}_{crop}$ & 23.3 \\
         & CLIP400M & CLIP & 14.0 \\
         & None & Random Guess & 12.5 \\
         & CLIP400M & $\text{CLIP}_{crop}$ & 11.2 \\

        \midrule
        \multirow{3}{*}{Generative} & COCO & $\oursRG$ & 31.2 \\
        & C4 + LAION-2b & Flamingo-9B & 12.5 \\
         & COCO & $\oursG$ & 0.0 \\

        % VecMap    & 28.4& 44.1 & 13.0 & 26.8 & 13.7&27.3\\
        \bottomrule[1pt]
        \end{tabular}
    }
    \vspace{-2mm}
    \caption{\small Zero shot transfer results of our Llama 2 backbone referring object classification model to new objects. \textit{crop} indicates cropping the image based on the bounding box annotation and using it as the input image. Flamingo is evaluated 8-shot (one example from each pathology).}
    \vspace{-4mm}
    \label{tab:zeroshot}

\end{table} 

\subsection{Referring Object Classification}

\paragraph{Unseen Object Classification} One of the benefits of our retrieval augmented system is its rapid generalization to unseen visual concepts. We estimate this capability by training on the COCO dataset and evaluating object classification on an unseen medical dataset which has drastically different types of images and limited training data. Table \ref{tab:zeroshot} illustrates the performance of our method on the CXR8 dataset in either a classification or generative setting. Even with as little as 20 examples per class in the medical retrieval set, $\oursR$ achieves competitive performance compared to domain-adapted models (i.e., BioMedCLIP), which we hypothesize is because of our region-level retrieval and in-distribution retrieval set. We also note that our generative approach $\oursRG$ can utilize the retrieved in-context examples and achieve similar performance to BioMedCLIP, despite only being trained on COCO images. Without retrieval, the generative model fails catastrophically with $0\%$ accuracy, and zero-shot CLIP achieves about the same performance as random guessing.

\begin{figure}[!t]
    \centering
    \includegraphics[width=\linewidth]{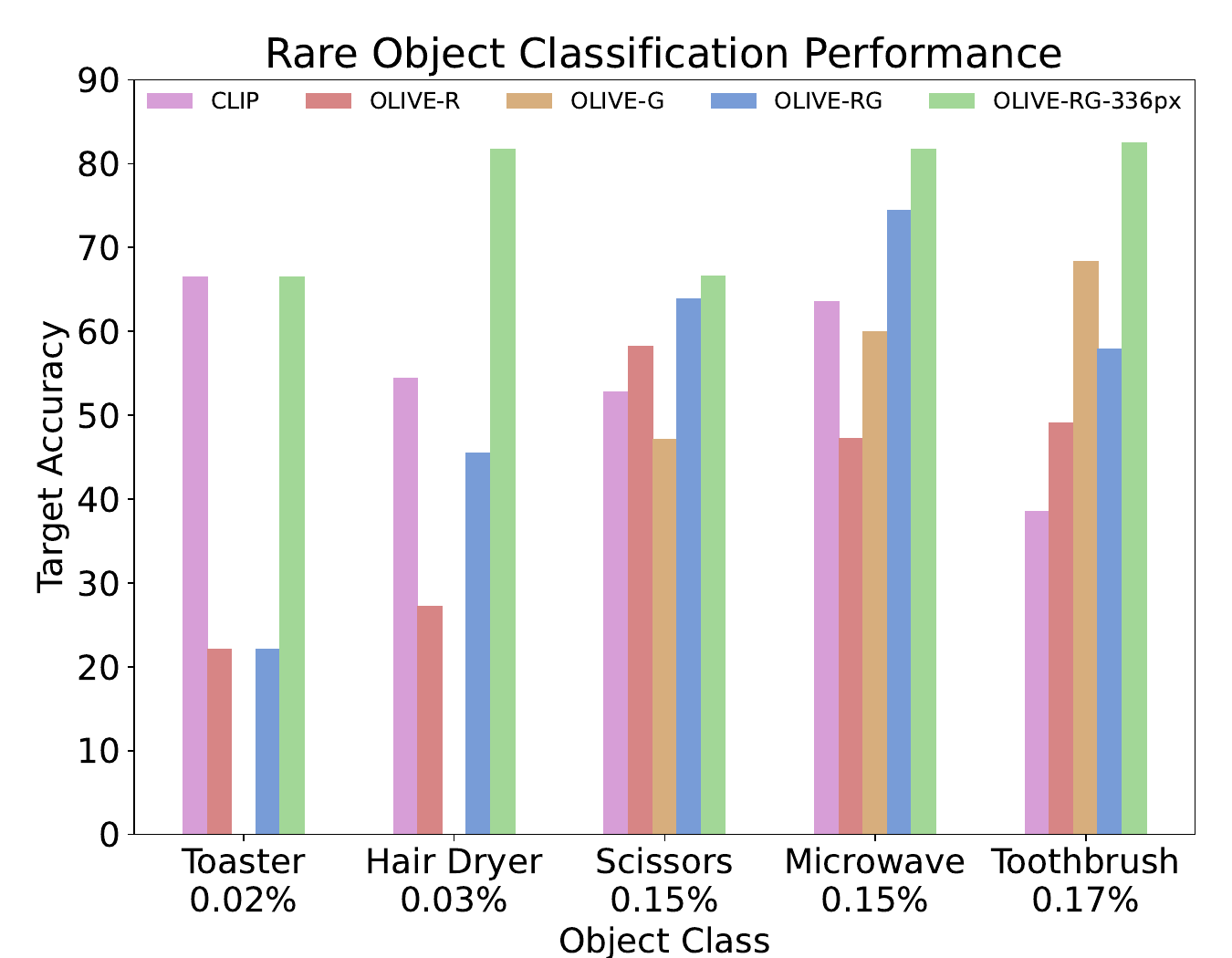}
    \vspace{-8mm}
    \caption{\small Classification accuracy on the most uncommon objects in the COCO dataset. We use a Llama 2 decoder backbone and numbers below the object classes indicate their proportion of the retrieval set. Combining retrieval with generative methods improves performance and increasing vision encoder resolution provides further gains.}
    \label{fig:rare_objects}
\end{figure}

 \paragraph{Rare Object Classification} We also investigate our model's performance on rare, but seen objects. Figure \ref{fig:rare_objects} shows our method's performance on the top 5 rarest classes in the COCO dataset. For $\oursG$ and $\oursRG$, we use a 224 pixel resolution visual encoder to match the CLIP visual encoder. $\oursG$ tends to have lower performances on the rare classes. However, when combining retrieval with parameterized methods in $\oursRG$ and $\oursRG$-336px, the performance on rare classes improves significantly, with $\oursRG$-336px performing better than CLIP on all rare classes. $\oursRG$ also achieves better performance on three out of five classes despite being trained on less data. Our model's overall performance can be found in Appendix \ref{sec:referring_object_classification}.

\subsection{Referring Expression Generation}

\paragraph{Captioning Unseen Objects} In addition to referring object classification, we investigate our model's ability to caption out-of-distribution objects. Figure \ref{fig:turtle} illustrates an example of asking our model to describe animals not seen during training. Without retrieval, $\oursG$ fails to describe the shark and turtle. However, after manually adding just 5 labeled objects of turtles and sharks to the existing retrieval set, $\oursRG$ accurately describes the object and provides supporting examples for its prediction. The label description for each object in the retrieval set is only the name of the animal, but the model generates additional characteristics in its description. Appendix \ref{sec:examples} shows more examples of zero-shot adaptation to unseen visual concepts in the object classification setting.

\begin{figure}
    \centering
    \includegraphics[width=0.48 \textwidth]{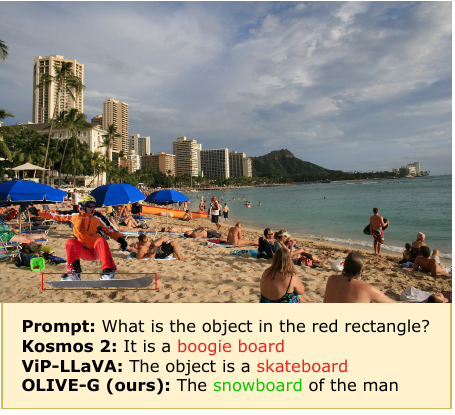}
    \caption{\small A challenging visual example in which the background of the image does not correspond with the query object. Methods which cross-attend to the whole image struggle to identify the snowboard, while our object representation enables more accurate description.}
    \label{fig:adversarial}
\end{figure}

\begin{figure*}
    \centering
    \includegraphics[width=1\textwidth]{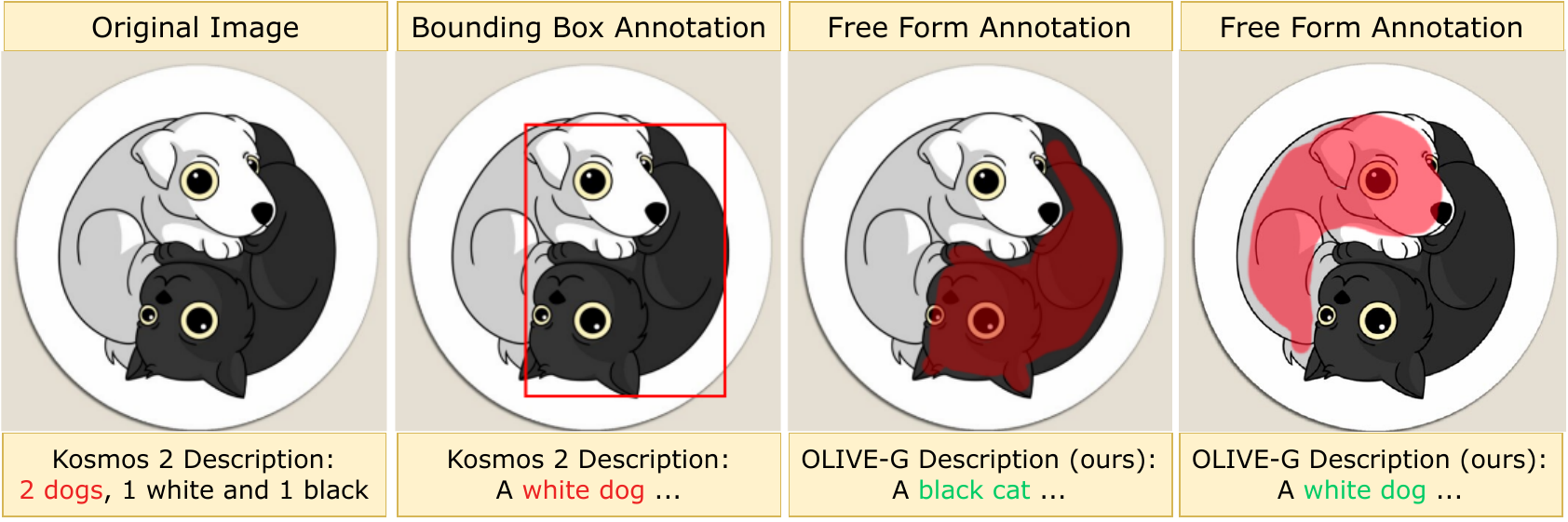}
    \caption{\small An illustration of the benefit of free-form visual input. Models that use the entire image or bounding box to refer to regions fail to describe the black cat, while $\oursG$ can use free-form annotation to identify the white dog and black cat.}
    \label{fig:yinyang}
\end{figure*}

\paragraph{Challenging Visual Context} To test the quality of the representations generated from our object encoder, we qualitatively evaluate our model prediction in adversarial visual contexts. Figure \ref{fig:yinyang} shows a white dog and a black cat in a ``yin-yang'' shape. We observe that free-form annotation allows for more precise user queries and object descriptions, and illustrates other properties such as scene content awareness and patch-level details as shown in Appendix \ref{sec:examples}. While many VLMs can accurately understand normal scenes, Figure \ref{fig:adversarial} illustrates an example in which an object-level representation may be necessary, with recent works struggling to caption the snowboarder on the beach. The detailed performance of our model on the refCOCOg captioning task can be found in Appendix \ref{sec:referring_expression_generation}.

\subsection{In-context Example Size}
\begin{figure}
    \centering
    \includegraphics[width=0.5\textwidth]{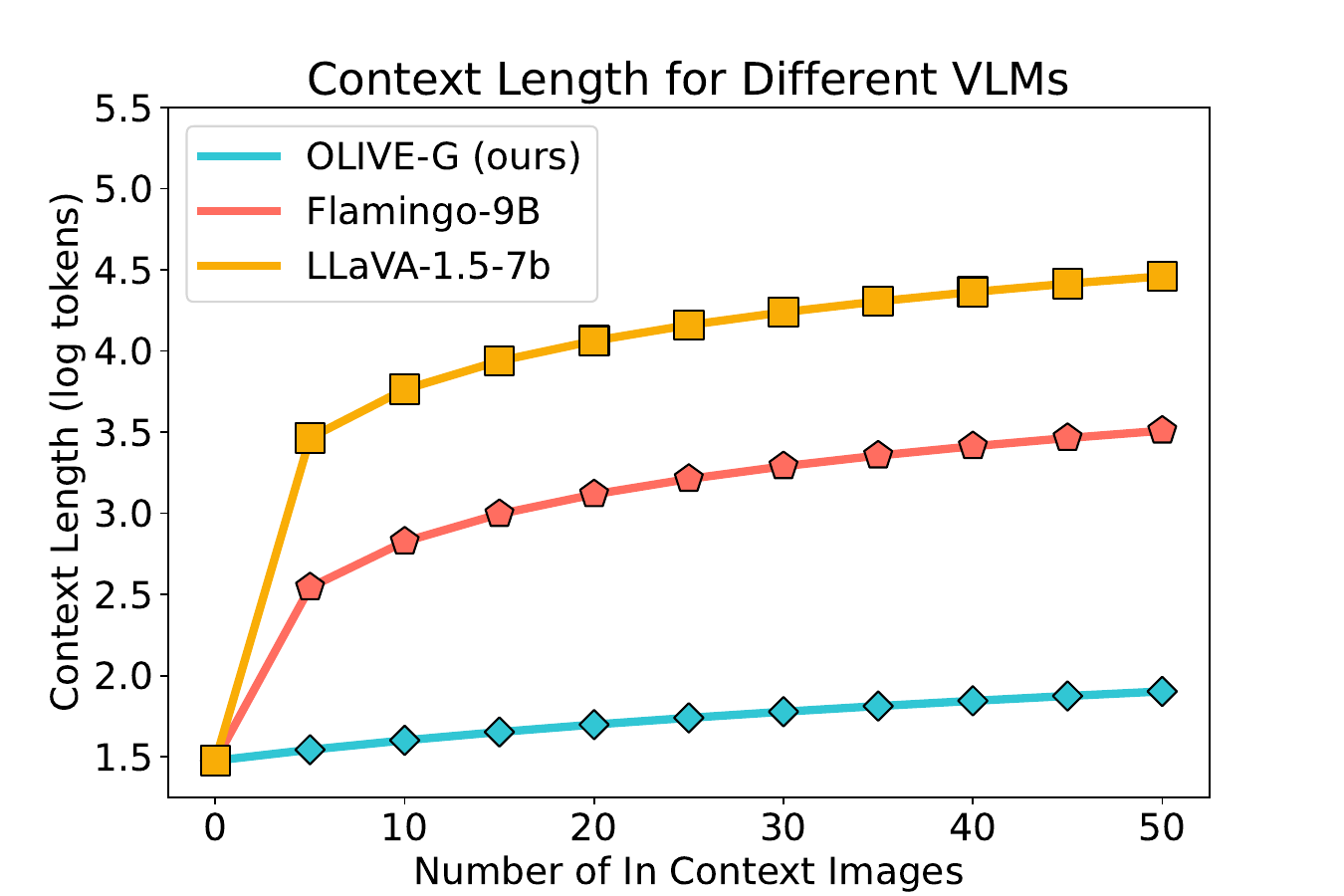}
    \caption{\small Context length of different VLMs when prompted with multimodal input. Models that represent images with many patch tokens or learned latents incur higher costs with more in-context examples.}
    \label{fig:context_length}
\end{figure}
Since our method omits image patch features and compresses object information into a single vector, it can process many objects from different images at once. In Figure \ref{fig:context_length}, we highlight the difference in context length for various methods when prompted with multimedia examples. We assume an average prompt length of 30 accompanying each in-context image example for all models. Even approaches designed for interleaved image-text data such as Flamingo insert multiple latent vectors for each image, incurring a higher cost than our approach.

\subsection{Sensitivity on Retrieval: Coverage and $k$}
\begin{figure}[!t]
    \centering
    \includegraphics[width=\linewidth]{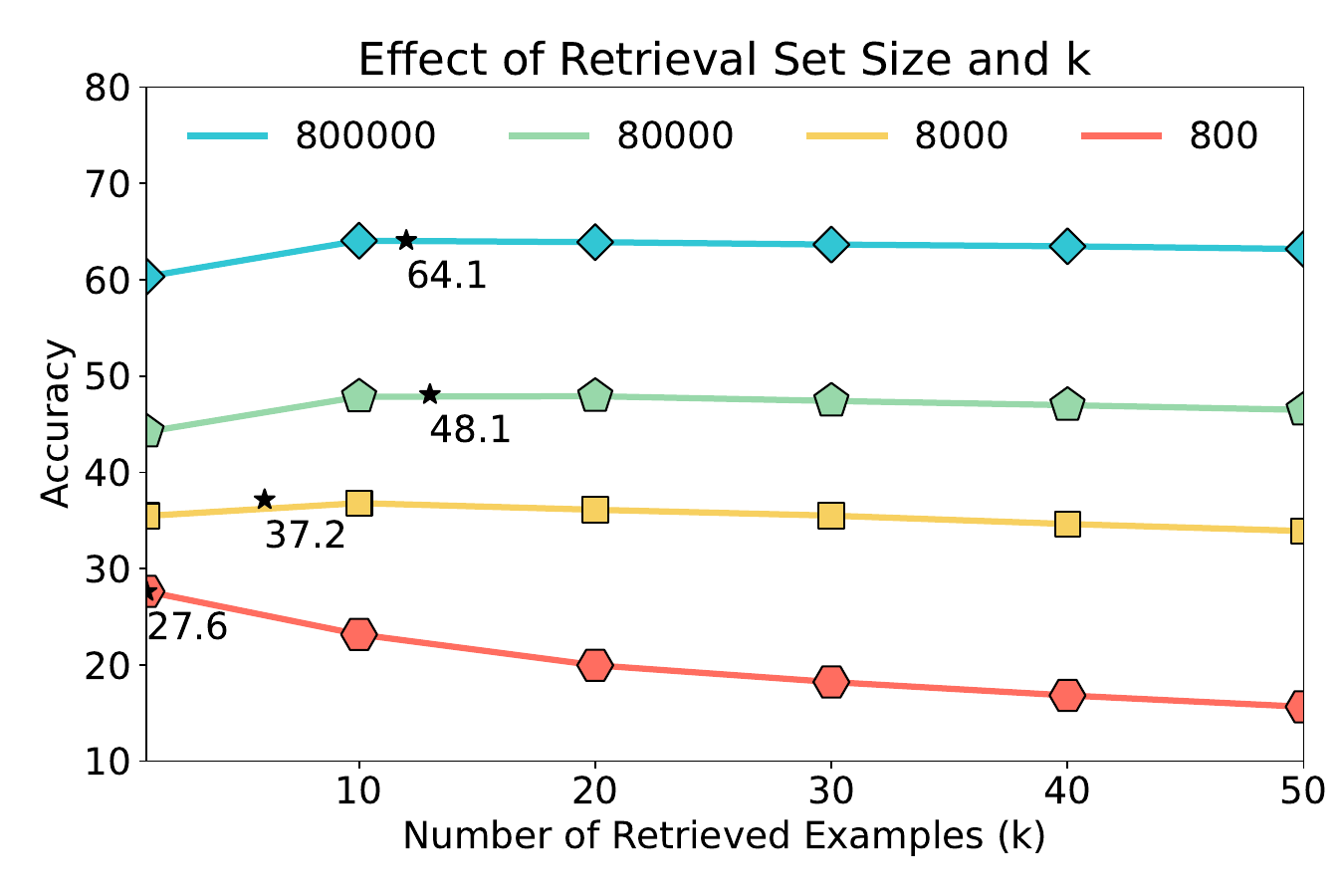}
    \vspace{-8mm}
    \caption{\small An illustration of how retrieval set size and choice of $k$ affects referring object classification performance. Numbers in the legend indicate the size of the retrieval set, and the stars are the highest accuracies achieved on their curves.}
    \label{fig:retrieval_set_size}
\end{figure}

In Figure \ref{fig:retrieval_set_size} we analyze the effect of changing the size of the object retrieval set as well as the number of retrieved examples, $k$. To thoroughly test various settings, we evaluate the retrieval-only based approach (\oursR) on the validation split of the COCO dataset using different sized subsets of the training data for retrieval. We ensure the retrieval set contains an equal amount of each object class when possible. Our results indicate that the optimal value of $k$ depends on the size of the retrieval set. With a small retrieval dataset (red), performance is lower and the optimal $k$ tends to be smaller. Larger retrieval sets (blue, green) benefit from retrieving more examples and have greater performance.

\subsection{Object Vector Visualization}
\begin{figure*}[!t]
    \centering
    \includegraphics[width=0.32\linewidth]{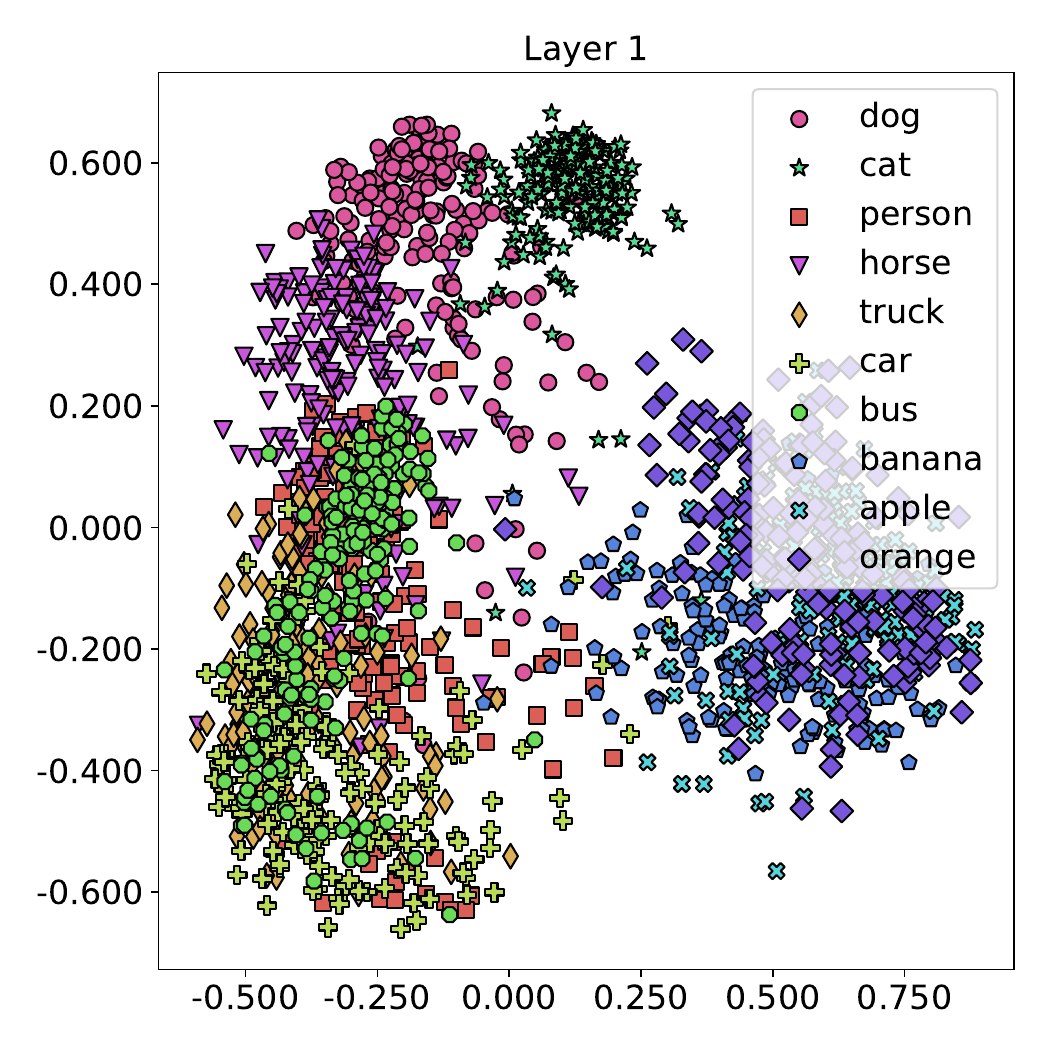}
    \includegraphics[width=0.32\linewidth]{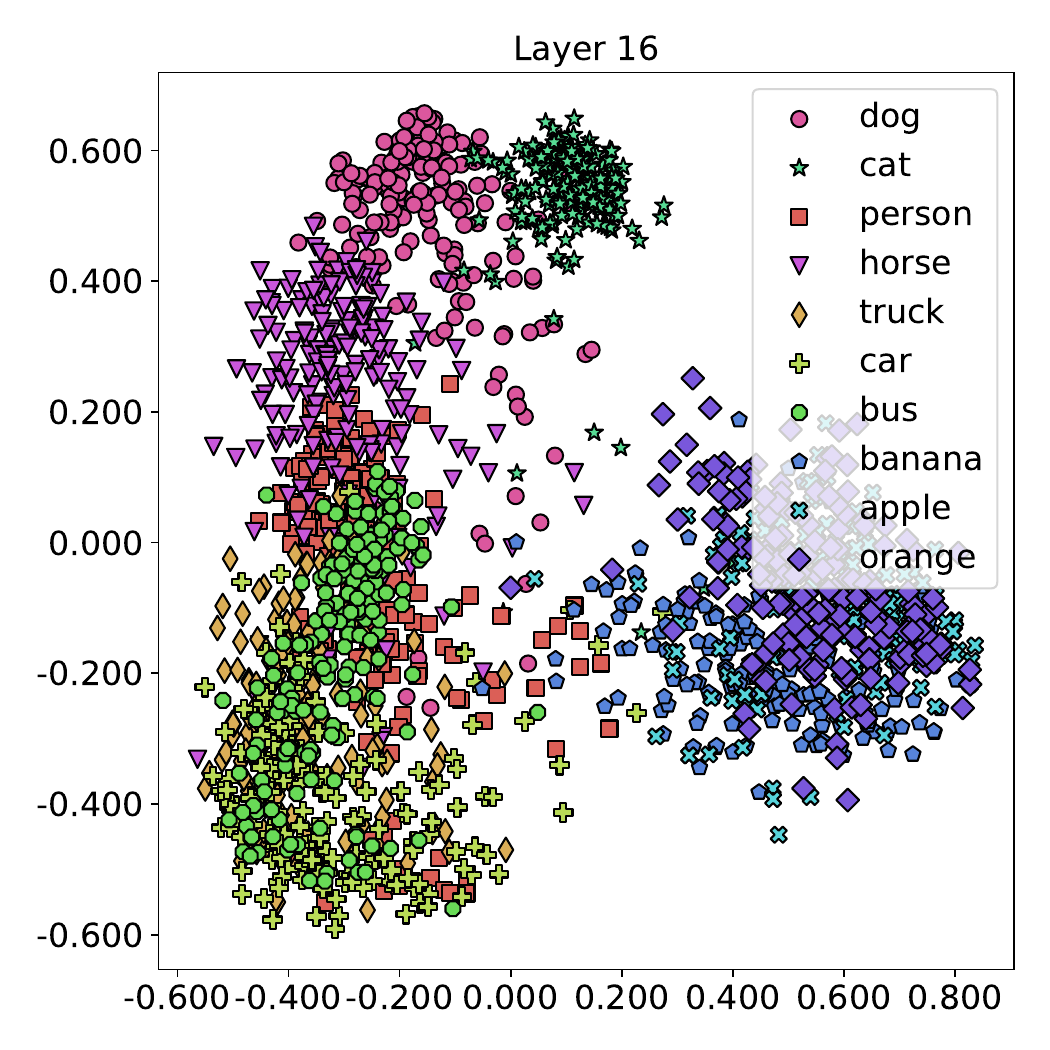}
    \includegraphics[width=0.32\linewidth]{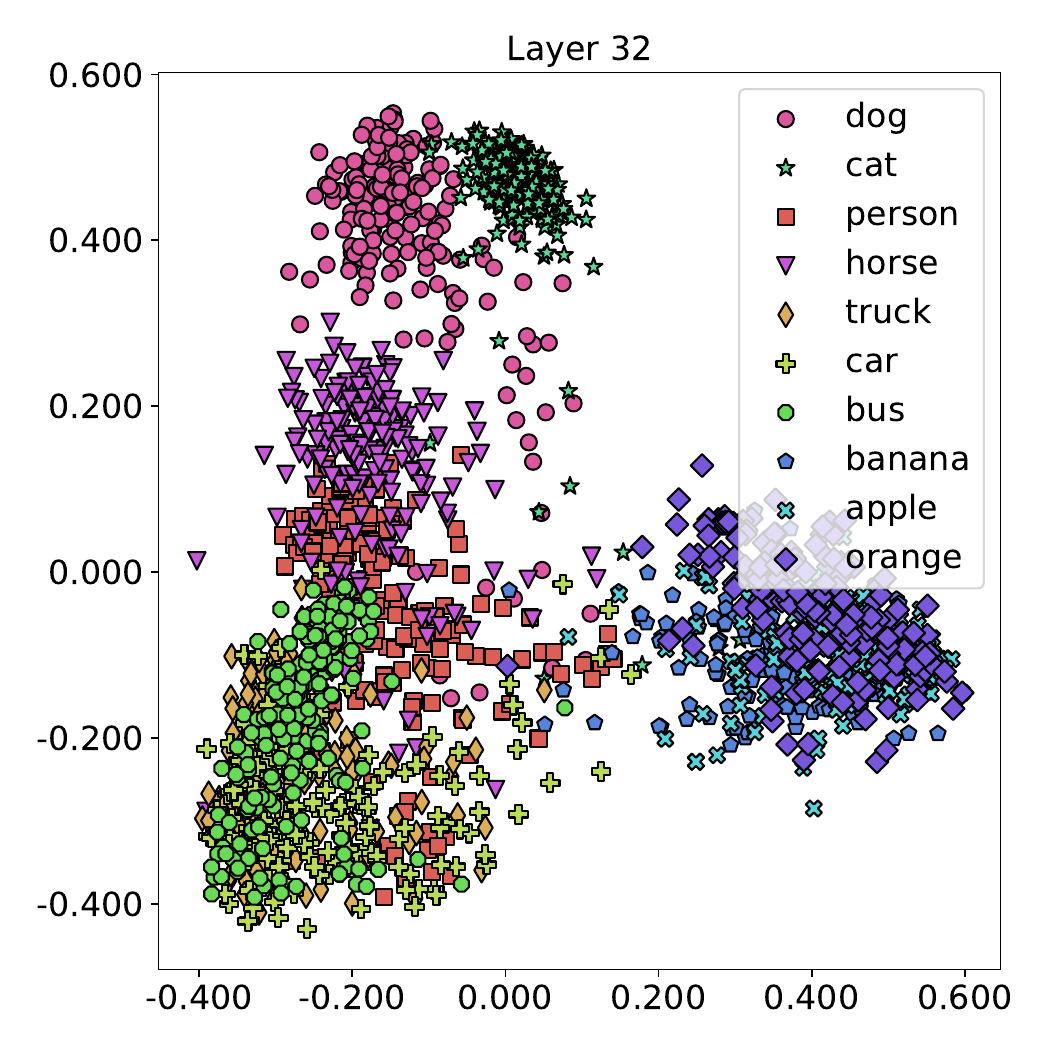}
    
    \caption{\small Visualization of the top 2 components when performing PCA decomposition on object vectors in different layers of the LLM decoder (Llama 2). We display 200 examples from each class. Across different layers, objects with similar semantics appear together in the plot, even though they appear in different visual contexts. The clusters are similar across all layers.}
    \label{fig:pca}
\end{figure*}

Having a single vector representation for each object allows for visualization using dimensionality reduction. In Figure \ref{fig:pca}, we perform principal component analysis (PCA) on the hidden states of object vectors at different layers in the LLM decoder. We plot 200 examples from each of 10 object categories and note several patterns. First, objects from the same class tend to appear together, even though they appear in different visual contexts. This suggests that the object encoder has semantic understanding of the visual concepts. Second, the object vectors naturally form hierarchical clusters where objects from the same super class such as vehicle, animal, or fruit have overlapping clusters. Lastly, the clustering appears similar across all layers, with only minor variations.

\section{Related Work}
\label{sec:related}

\paragraph{Grounding in Language and Vision}
A popular approach for aligning vision and language embeddings is contrastive learning methods such as CLIP and ALIGN \cite{li2021align}. However, these methods align the entire image representation, leading to poor reasoning on image details for downstream vision language tasks. RegionCLIP \cite{zhong2022regionclip} and GLIP \cite{li2022grounded} address this issue by proposing fine-grained alignment with region-text pairs during pretraining.  GLIPv2 \cite{zhang2022glipv2} further improves the pretraining and alignment by introducing localization, detection, and other tasks. Another recent popular approach involves training models on automatically curated region-level data from image-caption pairs \cite{peng2023kosmos}. Many other works focus on region-level alignment during pretraining for greater vision-language understanding \cite{you2023ferret, chen2023shikra, zeng2022x, zeng2021multi}. More generally, a recent study \cite{bugliarello2023measuring} shows that VLMs with fine-grained object-level pretraining such as X-VLM \cite{zeng2021multi} have better reasoning ability. Other works align vision and language using regularization or loss to create relation aware cross attention between modalities \cite{pandey-etal-2023-cross, ren2021learning}. 

\paragraph{Visual Resampling}
Visual resampling is a popular technique to compress long sequences of image features into a few rich vector representations. This is achieved by constructing a fixed amount of learnable vectors that attend to the visual features through cross-attention layers. Models such as BLIP \cite{li2023blip} first explore this idea to connect frozen vision features to LLMs efficiently by summarizing the content of the image. Other methods including X-Decoder \cite{zou2023generalized} or SEEM \cite{zou2023segment} use resampling to encode various types of prompts or intents which improve the LLM decoding ability. Additionally, works such as Flamingo \cite{alayrac2022flamingo} and Qwen-VL \cite{bai2023qwen} show that multiple images can be inserted in-context to the prompt by compressing image features with resamplers, enabling few shot capabilities. Our work visually resamples object representations for object-conditioned text generation, and only uses a single vector for the representation. This allows for more fine-grained reasoning and longer in-context prompting.

\paragraph{Retrieval Augmented VLMs}
In the text domain, learning to retrieve relevant documents to enhance the LLM query \cite{guu2020retrieval} has been explored extensively \cite{wang2023learning}. Recent VLM works follow a similar approach to retrieve multimodal documents to improve performance on knowledge-intensive tasks and improve generalization to rare situations. \citet{gao2022transform} summarizes visual content into natural language to use as a query for dense passage retrieval. MuRAG \cite{chen2022murag} proposes a multimodal image-text memory bank to help models answer challenging knowledge-based visual questions such as ``What shape is the pediment on top of the white house?" REVEAL \cite{hu2023reveal} and RA-VQA \cite{lin2022retrieval} learns a trainable multimodal retriever similar to REALM \cite{guu2020retrieval} during pretraining to fetch relevant documents to answer questions, achieving state of the art performance on datasets such as VQAv2 \cite{antol2015vqa} and OKVQA \cite{schwenk2022okvqa}. To the best of our knowledge, we are the first to integrate region-level retrieval with LLMs, in which the multimodal documents are indexed by object-level visual features.

% \paragraph{Transformers for Computer Vision}
% The transformer displays many interesting properties when applied to computer vision, such as robustness to random occlusion \cite{naseer2021intriguing}. Plain transformers have been shown to be effective backbones for object detection \cite{li2022exploring}, leading to powerful models such as Segment Anything (SAM) \cite{kirillov2023segment}. Furthermore, self-supervised training paradigms such as knowledge distillation with no labels (DINO) have shown new emerging capabilities such as semantic segmentation and high k-NN performance \cite{caron2021emerging}. Dinov2 \cite{oquab2023dinov2} recently improves the training process to be more stable and releases open-source models.

\section{Conclusion}
We present a simple approach to insert object level visual embeddings into large language model decoders, enabling object level reasoning with flexible prompt structure. Our object encoder compresses fine-grained region level information into a single vector, enabling in-context prompting with objects from multiple images and more efficient training and inference. In addition, we introduce the idea of region retrieval, which allows for precise queries free of image background noise and rapid generalization to rare and unseen objects with no parameter updates. We hope our method may help researchers design vision language models which can adapt to their needs by simply updating the retrieval set or object encoder, while also being responsive to varying user intents using LLM prompting techniques.
\label{sec:conclusion}

\section{Limitations}
\label{sec:limitations}
While our approach provides a flexible way for users to supply object-level prompts, it does not output bounding boxes or other region-level grounding. This may be addressed in future research by further finetuning on region-level instruction tuning data as done in FERRET \cite{you2023ferret}, GLAMM \cite{rasheed2023glamm}, and other region-level VLM pretraining. At the moment, we also do not explore generic image tasks such as VQA or image captioning. However, a potential solution is to use our object encoder to connect to existing VLMs (e.g. LLaVA) which excel at these tasks. Lastly, our results in the retrieval setting depend on the quality of the retrieved examples. Curating a high-quality retrieval set at the object-level can be challenging. However, existing tools such as GLIPv2 \cite{zhang2022glipv2} allows for semi-automatic generation of region-level data as used in KOSMOS-2 \cite{peng2023kosmos} in developing the GRIT dataset.
\section{Ethical Considerations}
\label{sec:ethics}
\paragraph{Biases From Pretrained LLMs} Since our model uses existing pretrained LLMs such as Llama 2 or GPT2, it may inherent some of the social biases or toxicity acquired during their pretraining stages. While Llama 2 undergoes extensive alignment to ideal human values through reinforcement learning from human feedback (RLHF) \cite{griffith2013policy}, some of these toxic behaviors may still be present in the morally aligned model. We make sure to only use images of common objects in the COCO dataset, which do not contain any of these biases or violent scenes to the best of our knowledge. Nevertheless, further testing to ensure the impartiality of the model may be necessary before deploying in widespread technologies.

\paragraph{Domain Adaptation}
Some of our experiments involve evaluating our model in a data-scarce domain in a zero shot manner with in-context prompting. While this is a promising direction for efficient domain adaptation, users should take caution in directly using model prediction, as this is a challenging task due to distribution shift. We encourage human-in-the-loop interaction to sanity check the outputs. Different from other ICL prompting methods, we provide retrieved examples and similarity scores which can help determine the trustworthiness of the model prediction, which may be valuable for high-risk domains such as medicine. 
\section*{Acknowledgement}
Ossowski and Hu are supported by the Wisconsin Alumni Research Foundation and the National Institute Of Biomedical Imaging And Bioengineering of the National Institutes of Health under Award Number R01EB033782. The content is solely the responsibility of the authors and does not necessarily represent the official views of the National Institutes of Health.

% Entries for the entire Anthology, followed by custom entries
\bibliography{anthology,custom}
\bibliographystyle{acl_natbib}

\clearpage
\appendix

\section{Appendix}

\label{sec:prompts}
\begin{table*}[!htbp]
   \resizebox{\linewidth}{!}
    {
    \centering
    \scriptsize
        \begin{tabular}{c@{\hspace{1ex}}c@{\hspace{1ex}}c}
        % \hline
        \toprule[1pt]
        % \multirow{2}{*}{Method} & \multicolumn{2}{c}{En$\rightarrow$Ru} & \multicolumn{2}{c}{En$\rightarrow$It} & \multicolumn{2}{c}{En$\rightarrow$ES} & \multicolumn{2}{c}{En$\rightarrow$Fr} & \multicolumn{2}{c}{En$\rightarrow$DE}\\
        Task & ICL Prompt for Retrieved Examples & Vanilla Prompts \\
        
         % \hline
         \midrule 
        \makecell{Object \\ Classification} & \makecell{You are a helpful vision assistant trained to help people analyze images. \\ The top \texttt{[k]} related objects are: \\
        \texttt{[obj]} is a \texttt{[label]} with confidence \texttt{[score]} \\
        \texttt{[obj]} is a \texttt{[label]} with confidence \texttt{[score]} \\
        \vdots \\
         \texttt{[vanilla prompt]}} & \makecell{\texttt{[obj]} What is this? Answer in 1-2 words \\ \texttt{[obj]} What is this object? Answer with a short word or phrase.\\ \texttt{[obj]} Identify this object. \\ Here is an object \texttt{[obj]}. What is this? Answer with a short word or phrase.} \\

        \midrule
        \makecell{Region \\ Description} & \makecell{You are a helpful vision assistant trained to help people analyze images. \\ The top \texttt{[k]} related objects are: \\
        \texttt{[obj]} is a \texttt{[label]} with confidence \texttt{[score]} \\
        \texttt{[obj]} is a \texttt{[label]} with confidence \texttt{[score]} \\
        \vdots \\
        \texttt{[vanilla prompt]}} & \makecell{\texttt{[obj]} Briefly describe this image region. \\
        \texttt{[obj]} Describe this part of the image. \\
        \texttt{[obj]} Share some details about about what's happening here in the image. \\
        \texttt{[obj]} Break down what you see in this particular part of the picture. \\
        \texttt{[obj]} Describe what you notice in this area of the picture. \\} \\
        % \cline{3-5}
        % {}& $\rightarrow$  & {$\leftarrow$} & $\rightarrow$ & $\leftarrow$ & $\rightarrow$ & $\leftarrow$ & $\rightarrow$ & $\leftarrow$ & $\rightarrow$ & $\leftarrow$ \\
    
        % & RepsNet \cite{tanwani2022repsnet} & - & - & - & - & - & 81.1 \\
        % & Medical Knowledge Pre-training \cite{chen2022align} & 81.9 & 91.4 & 85.6 & 67.6 & 86.8 & 79.2\\
        
        \bottomrule[1pt]
        \end{tabular}
        }
    \vspace{-3mm}
    \caption{ Collection of the prompts we use to guide the LLM decoder generation. For a given task, we sample one of the vanilla prompts uniformly at random. Text in brackets indicates variables whose value is dynamically filled in.}
    \label{tab:prompts}

\end{table*}
Table \ref{tab:prompts} contains all the prompts we use to instruct the LLM decoder.

\section{Qualitative Examples}
\label{sec:examples}

\begin{figure*}
    \begin{center}
        
\includegraphics[width=\textwidth]{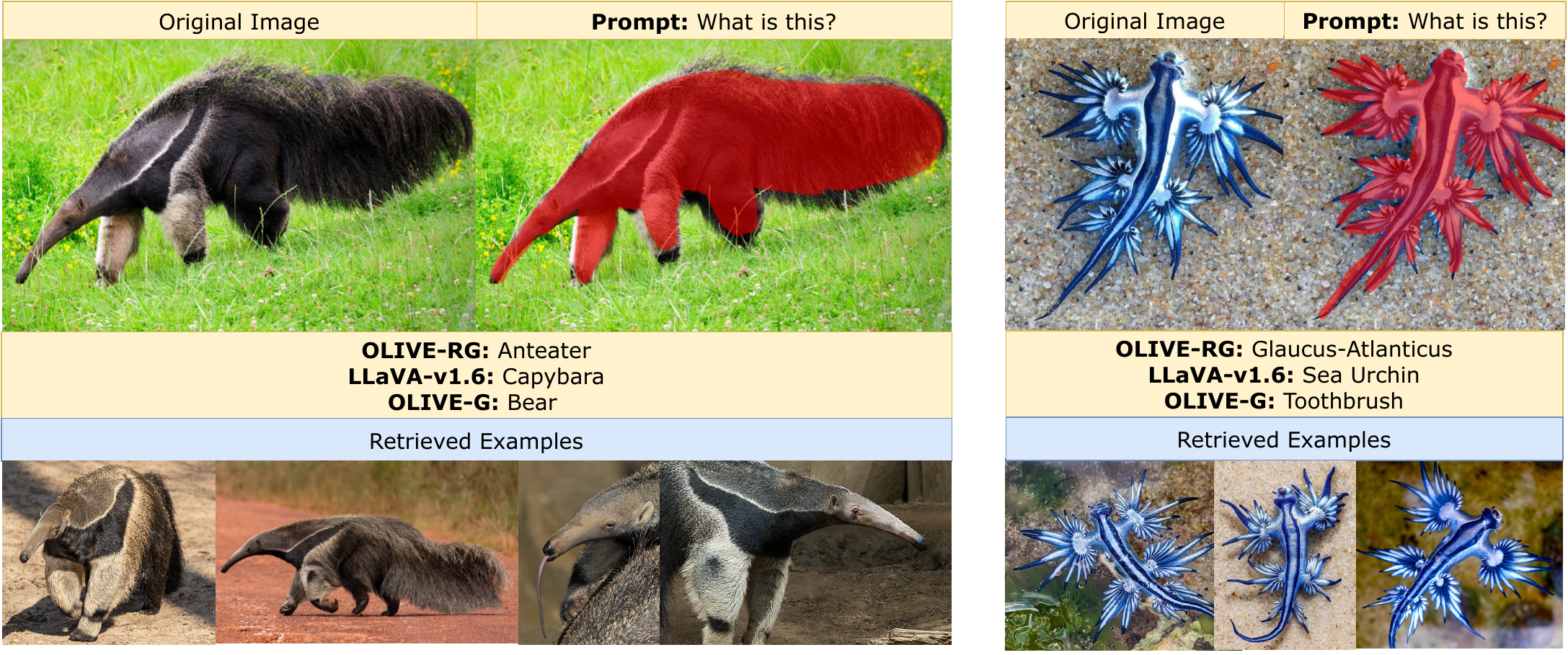}
    
    \caption{Examples of rapid adaptation to unseen visual concepts during training. Non-retrieval-based methods such as LLaVA often fail to generalize, instead predicting animals seen during pre-training.}
    \label{fig:retrieval_examples}
    \end{center}

\end{figure*}

\begin{figure*}
    \begin{center}
        
    \includegraphics[width=\textwidth]{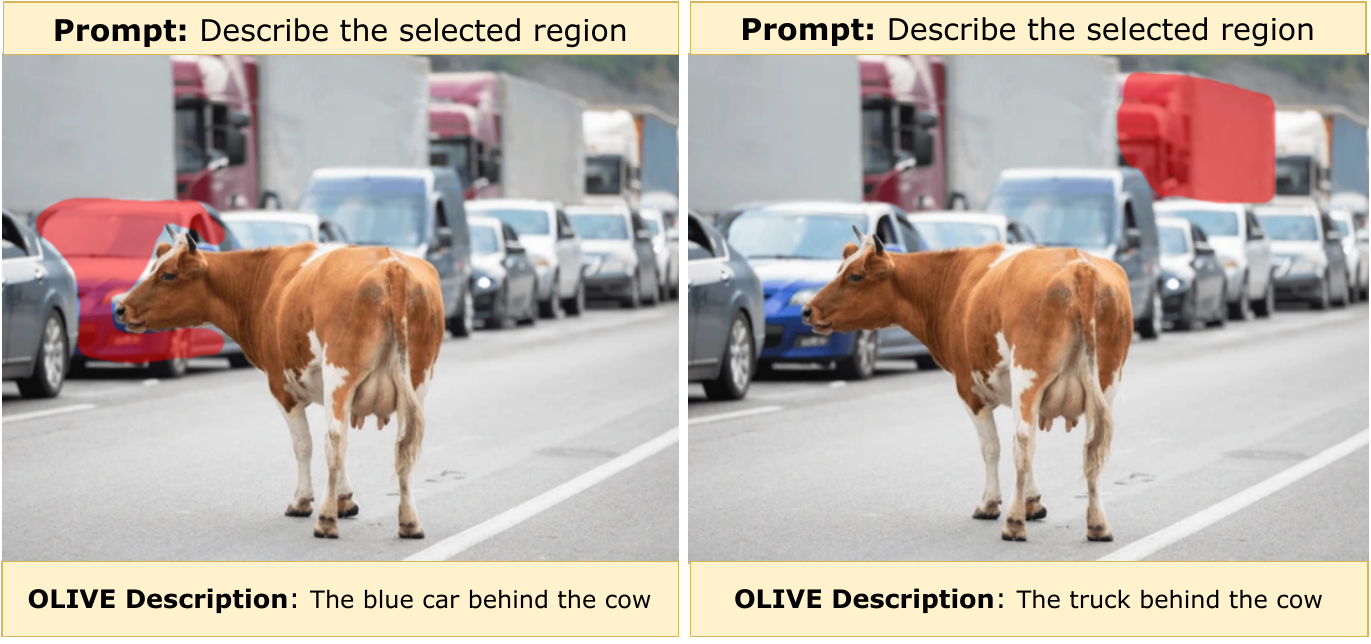}
    \caption{Example of scene content awareness of the object embeddings. Although neither selection includes any part of the cow, the model can still mention the cow in its description.}
    \label{fig:cowpic}
    \end{center}

\end{figure*}

\begin{figure*}
    \begin{center}
        
    \includegraphics[width=\textwidth]{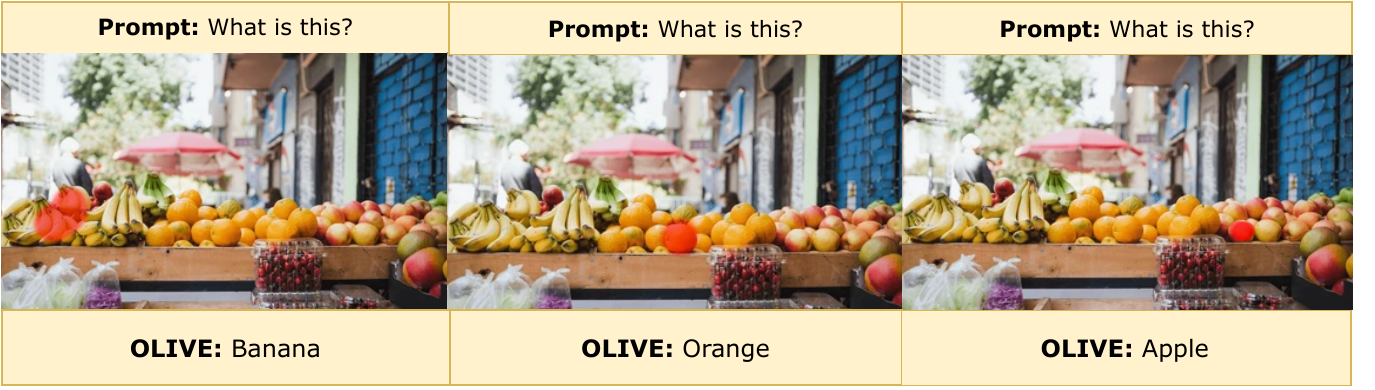}
    \caption{Object classification on small objects.}
    \label{fig:fruits}
    \end{center}

\end{figure*}

\begin{figure*}
    \begin{center}
        
    \includegraphics[width=\textwidth]{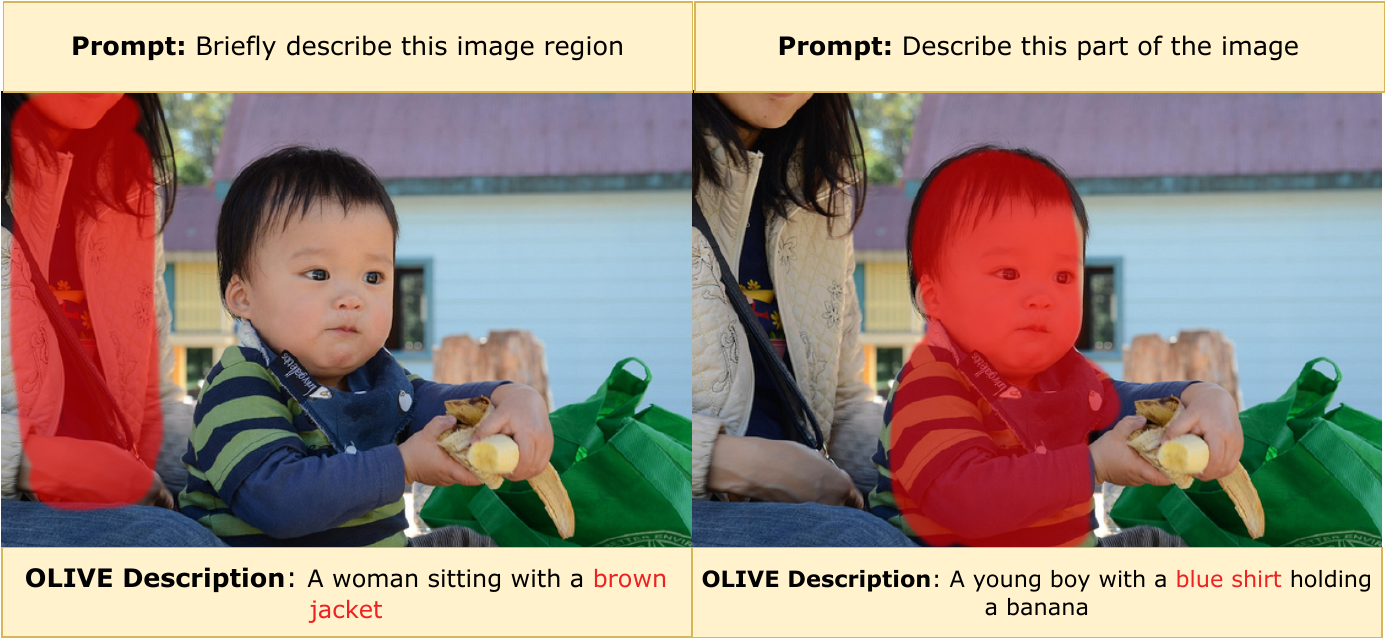}
    \caption{Failure cases of our model. The model sometimes struggles with occluded/partially missing objects and may mix up some fine-grained details about objects.}
    \label{fig:failure}
    \end{center}

\end{figure*}

Here we include several selected examples showcasing the strengths and weaknesses of our approach.

\paragraph{Visual Concept Generalization}
In Figure \ref{fig:retrieval_examples} we demonstrate more examples of rapid generalization to new visual concepts. Many existing methods confidently predict concepts from their pretraining, while ours can predict new concepts on the fly.

\paragraph{Scene Content Awareness}
Even though our object representation involves masking out image patch features from other parts of the image, we have observed that the object vector still contains information about its surroundings. Figure \ref{fig:cowpic} illustrates this phenomenon, where $\ours$ can include the cow in its description, despite not including any image patches corresponding to the cow in the user selection.

\paragraph{Patch level Detail}
Our method also can identify and describe small objects at the patch level. Figure \ref{fig:fruits} shows an example of object classification on smaller objects.

\paragraph{Describing Partially Visible Objects}
We notice that our model can make mistakes when describing occluded or partially visible objects as seen in Figure \ref{fig:failure}. We hypothesize that the training data of refCOCOg does not include these kinds of image regions, which also limits its availability in retrieval data. This may be addressed with larger-scale pre-training on data such as GRIT which likely includes more occluded objects. 

\paragraph{Errors in Detailed Description}
While our model can identify the object most of the time, it sometimes gets minor details incorrect. For example, the colors of a shirt or other piece of clothing are seen in Figure \ref{fig:failure}. This may be due to the extreme compression we learn into a single vector. Future work may consider visually resampling the object features into more than just 1 latent vector for detailed captioning, but still use the single vector representation for retrieval.

\section{Dataset Information}
\begin{table*}[!htbp]
   \resizebox{\linewidth}{!}
    {
    \centering
    \scriptsize
        \begin{tabular}{cccccc}
        
        \toprule[1pt]
        % \multirow{2}{*}{Method} & \multicolumn{2}{c}{En$\rightarrow$Ru} & \multicolumn{2}{c}{En$\rightarrow$It} & \multicolumn{2}{c}{En$\rightarrow$ES} & \multicolumn{2}{c}{En$\rightarrow$Fr} & \multicolumn{2}{c}{En$\rightarrow$DE}\\
        Dataset & Train Split & Validation Split & Retrieval Set Train Split & Retrieval Set Test Split & Number of Classes \\
        \midrule 
        
        COCO  & 849586 & 36320 & 849586 &  849586 & 80 \\

         refCOCOg & 44822 & 5000 & 849586 & 849586 & - \\

        CXR8 & - & 824 & - & 160 & 8 \\

        % \cline{3-5}
        % {}& $\rightarrow$  & {$\leftarrow$} & $\rightarrow$ & $\leftarrow$ & $\rightarrow$ & $\leftarrow$ & $\rightarrow$ & $\leftarrow$ & $\rightarrow$ & $\leftarrow$ \\
    
        % & RepsNet \cite{tanwani2022repsnet} & - & - & - & - & - & 81.1 \\
        % & Medical Knowledge Pre-training \cite{chen2022align} & 81.9 & 91.4 & 85.6 & 67.6 & 86.8 & 79.2\\
        
        \bottomrule[1pt]
        \end{tabular}
        }
    \vspace{-3mm}
    \caption{More details about the datasets and splits we use in our experiments. For COCO and refCOCOg, we use the train split of the COCO dataset as the retrieval data. We select the first 20 examples of each pathology as the retrieval set for the zero shot evaluation on CXR8.}
    \label{tab:datasets}

\end{table*}
\label{sec:datasets}
Table \ref{tab:datasets} provides more details on the dataset splits used in our training and evaluation. Our COCO train and validation splits are slightly smaller than normal because of our approach of using segmentation masks. We decide to omit some excessively small segmentations which account for less than 1\% of the data. For tasks that require training (COCO and refCOCOg), we use the train split of the COCO object detection dataset as our retrieval data. We make sure to omit the closest match when training object detection on COCO with retrieval to avoid label leakage. We also confirm that no images are repeated in the validation split from the training split for both datasets.

\section{Referring Object Classification}
\begin{table}[!htbp]
   \resizebox{\linewidth}{!}
    {
    \centering
    \scriptsize
        \begin{tabular}{clcccc}
        \toprule[1pt]
        % \multirow{2}{*}{Method} & \multicolumn{2}{c}{En$\rightarrow$Ru} & \multicolumn{2}{c}{En$\rightarrow$It} & \multicolumn{2}{c}{En$\rightarrow$ES} & \multicolumn{2}{c}{En$\rightarrow$Fr} & \multicolumn{2}{c}{En$\rightarrow$DE}\\
        % & & & \multicolumn{2}{c}{Performance} \\ 
         Method Type & Method & Accuracy & mAP \\

        % \cline{3-5}
        % {}& $\rightarrow$  & {$\leftarrow$} & $\rightarrow$ & $\leftarrow$ & $\rightarrow$ & $\leftarrow$ & $\rightarrow$ & $\leftarrow$ & $\rightarrow$ & $\leftarrow$ \\
        % \hline
        \midrule 
        \multirow{4}{*}{Classification} 
        & $\oursR$  & \textbf{64.1} & 40.5 \\
        & CLIP \texttt{ViT-L/14}& 40.9 & 45.1 \\
        & RegionCLIP \texttt{RN50} & - & \textbf{61.4} \\
        &  OVR & - & 44.5  \\
        % & RepsNet \cite{tanwani2022repsnet} & - & - & - & - & - & 81.1 \\
        % & Medical Knowledge Pre-training \cite{chen2022align} & 81.9 & 91.4 & 85.6 & 67.6 & 86.8 & 79.2\\
        \midrule 
         \multirow{4}{*}{Generative}

         & $\oursG$ (GPT2) & 76.6 & \textbf{60.4}  \\
         & $\oursG$ (Llama 2)  & \textbf{76.8} & 60.3 \\

         % & Llama 2  & 73.3 & 67.7  \\

         & $\oursRG$ (GPT2) & 74.8 & 57.5  \\
         & $\oursRG$ (Llama 2)  & 74.1 & 56.2 \\
        \bottomrule[1pt]
        \end{tabular}
        }
    \vspace{-3mm}
    \caption{ \small Performances on the referring object classification task with different levels of context on the COCO dataset. For each method type, the highest values for each metric are bolded. }
    \label{tab:overall}

\end{table}
\label{sec:referring_object_classification}
This task requires the LLM to predict the object class label given a ground truth input annotation (e.g. bounding box, segmentation, etc). We follow a similar evaluation protocol used in \cite{ zhong2022regionclip} and \cite{zareian2021open}, in which the ground truth annotation is supplied to avoid localization error. Table \ref{tab:overall} shows the overall referring object classification accuracy and mAP \footnote{To simplify the calculation, we assigned a confidence score of 1 to each prediction. Reported mAP may be lower than the true value when using more accurate probabilities.} for our methods. We observe several findings. First, although retrieved examples help with domain adaption and rare objects, it does not improve the overall in-domain performance. Second, both the LLama 2 and GPT2 baseline have similar performances on the task, suggesting that even smaller models can learn vision-language grounding. Lastly, even our retrieval-only baseline, which requires no training, has better accuracy than some parameterized methods such as CLIP.

\begin{figure*}[!ht]
    \centering
    \includegraphics[width=\textwidth]{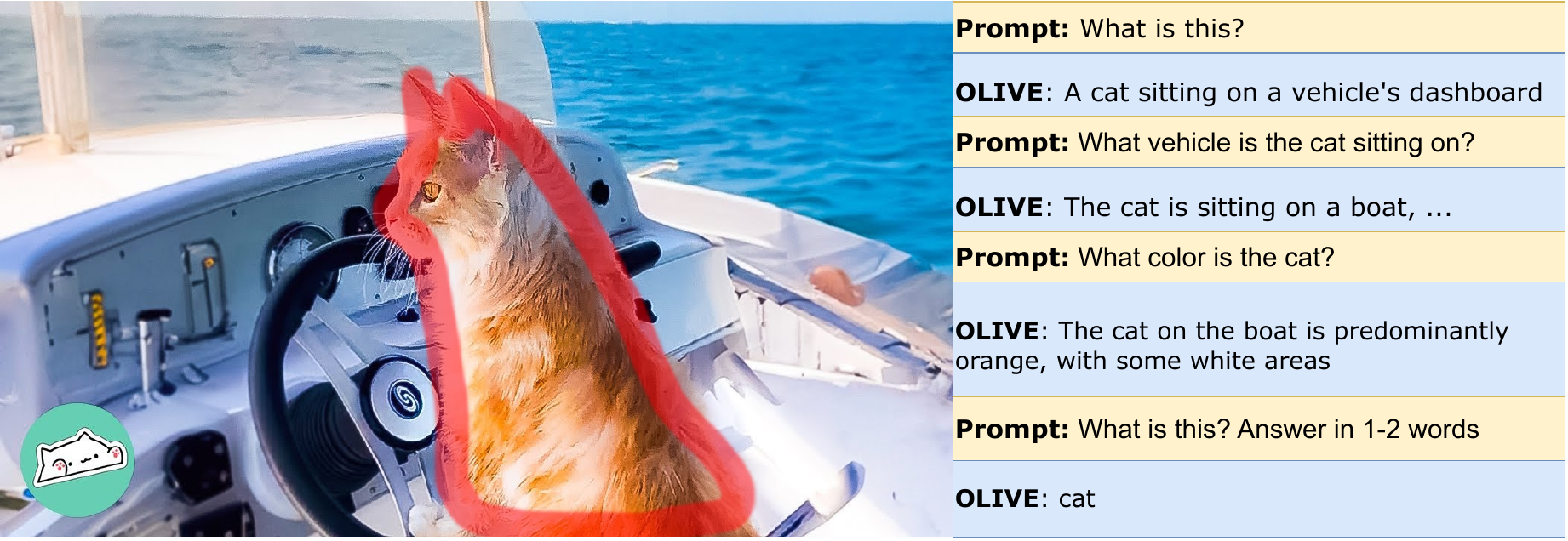}
    \caption{When trained on multiple tasks and object instruction following data, OLIVE is able to respond to varying user intents.}
    \label{fig:instruction_following}
\end{figure*}

\section{Multi-Task Model}
\label{sec:multi_task_model}

We also explore the possibility of training a multi-task model using a similar curriculum learning strategy to LLaVA \cite{liu2023visual}. We first train the model on the referring object classification task to perform the object-word level alignment. The model is then trained on the referring expression generation task, and finally on an object instruction following dataset \cite{cai2023making} with many different tasks. For each stage of training, we formulate the task in an instruction-following manner through the prompts in Table \ref{tab:prompts}. This allows the model to be responsive to many different user intents (Figure \ref{fig:instruction_following})

\section{Referring Expression Generation}
\begin{table}[!htbp]
   \resizebox{\linewidth}{!}
    {
    \centering
    \scriptsize
        \begin{tabular}{ccc}
        % \hline
        \toprule[1pt]
        % \multirow{2}{*}{Method} & \multicolumn{2}{c}{En$\rightarrow$Ru} & \multicolumn{2}{c}{En$\rightarrow$It} & \multicolumn{2}{c}{En$\rightarrow$ES} & \multicolumn{2}{c}{En$\rightarrow$Fr} & \multicolumn{2}{c}{En$\rightarrow$DE}\\
        % & & & \multicolumn{2}{c}{Performance} \\ 
         Method &  METEOR & CIDEr \\

        % \cline{3-5}
        % {}& $\rightarrow$  & {$\leftarrow$} & $\rightarrow$ & $\leftarrow$ & $\rightarrow$ & $\leftarrow$ & $\rightarrow$ & $\leftarrow$ & $\rightarrow$ & $\leftarrow$ \\
        % \hline
        \midrule 
         % $\ours_{gen}$ &  13.3 & 56.2 \\
         % $\ours_{gen\_lora}$ & 16.5 & 64.0\\
         $\oursG$ (Llama 2) & 16.5 & 64.0 \\
          $\oursRG$ (Llama 2) & 16.6 & 67.7\\
          $\oursG$ (GPT2) & 16.4 & 70.9 \\
          $\oursRG$ (GPT2) & \textbf{17.0} & 75.0 \\
          SLR \cite{yu2017joint} & 15.4 & 59.2\\
          SLR+Rerank \cite{yu2017joint} & 15.9 & 66.2 \\
         GLAMM \cite{rasheed2023glamm} & 16.2 & \textbf{105.0} \\
         GRIT \cite{wu2022grit} & 15.2 & 71.6 \\
         Kosmos 2 (zero shot) & 12.2 & 60.3 \\
         Kosmos 2 (fewshot k = 2) & 13.8 & 62.2 \\
         Kosmos 2 (fewshot k = 4) & 14.1 & 62.2 \\
         Flamingo-9B (zero shot) & 9.2 & 34.3\\
         Flamingo-9B (fewshot k = 2) & 10.2 & 36.2 \\
         Flamingo-9B (fewshot k = 4) & 12.3 & 39.6 \\
        \bottomrule[1pt]
        \end{tabular}
        }
    \vspace{-3mm}
    \caption{ \small Referring expression generation on the refCOCOg validation split. Our approach has competitive perfomance compared to other notable methods which also offer multimodal in-context prompting.}
    \label{tab:captioning}

\end{table}
\label{sec:referring_expression_generation}

We study our model's overall performance on referring expression generation by quantitatively evaluating our model on the RefCOCOg validation set shown in Table \ref{tab:captioning}. Several findings can be observed. First, including retrieved multimodal documents results in slightly better performance. Second, the size of the LLM can be modified without much performance change, with GPT2 performing slightly better than Llama 2. Third, having global image context contained in the object representation is important, as methods that crop the image region (e.g. Flamingo) perform worse.

\section{Training Hyperparameters}
We provide the detailed training hyperparameters in Table~\ref{tab:hyperparameters}.
\begin{table}[!htbp]
   \resizebox{\linewidth}{!}{
    \centering
    % \scriptsize
        \begin{tabular}{ccc}
        % \hline
        \toprule[1pt]
        % \multirow{2}{*}{Method} & \multicolumn{2}{c}{En$\rightarrow$Ru} & \multicolumn{2}{c}{En$\rightarrow$It} & \multicolumn{2}{c}{En$\rightarrow$ES} & \multicolumn{2}{c}{En$\rightarrow$Fr} & \multicolumn{2}{c}{En$\rightarrow$DE}\\
        % Hyperparameter & Referring Object Classification & Referring Expression Generation \\
        Hyperparameter & Classification & Generation \\

        \midrule 
        Epochs & 1 & 5 \\
        Batch Size & 4 & 4 \\
        Training Steps & $\sim$ 200,000 & $\sim$ 56,030 \\
        Learning Rate & 2e-5 & 2e-5 \\
        Optimizer & Adam & Adam \\
        GPU Used & GTX 3090 & GTX 3090 \\
        Train Time (hours) & 24 & 7.5 \\

        % \cline{3-5}
        % {}& $\rightarrow$  & {$\leftarrow$} & $\rightarrow$ & $\leftarrow$ & $\rightarrow$ & $\leftarrow$ & $\rightarrow$ & $\leftarrow$ & $\rightarrow$ & $\leftarrow$ \\
    
        % & RepsNet \cite{tanwani2022repsnet} & - & - & - & - & - & 81.1 \\
        % & Medical Knowledge Pre-training \cite{chen2022align} & 81.9 & 91.4 & 85.6 & 67.6 & 86.8 & 79.2\\
        
        \bottomrule[1pt]
        \end{tabular}
        }
    \vspace{-3mm}
    \caption{ Details about the hyperparameters we use for (1) Referring Object Classification (\textbf{Classification}) and (2) Referring Expression Generation (\textbf{Generation}) in our experiments.}
    \label{tab:hyperparameters}

\end{table}

% \section{Vision Encoder Ablation}
% \input{tables/overall.tex}
% \label{sec:vision_encoder}
% We tried different vision encoders during our experiments, and ultimately found that patch

\end{document}